\documentclass[11pt, a4paper, logo, onecolumn, copyright]{googledeepmind}

\usepackage[authoryear, sort&compress, round]{natbib}
\usepackage{xspace}  
\usepackage{siunitx}  
\usepackage[capitalise,noabbrev]{cleveref}
\usepackage{placeins}
\usepackage{subfig}

\title{Skillful joint probabilistic weather forecasting from marginals}

\correspondingauthor{{\{ferran,alvarosg,peterbattaglia\}@google.com}}

\author[1*]{Ferran Alet}
\author[1*]{Ilan Price}
\author[1*]{Andrew El-Kadi}
\author[1*]{Dominic Masters}
\author[1]{Stratis Markou}
\author[1]{Tom R. Andersson}
\author[1]{Jacklynn Stott}
\author[1]{Remi Lam}
\author[1]{Matthew Willson}
\author[1*]{Alvaro Sanchez-Gonzalez}
\author[1]{Peter Battaglia}

\affil[*]{Equal contributions}
\affil[1]{Google DeepMind}

\newcommand{\ourmodel}{FGN\xspace}
\newcommand{\gc}{GraphCast\xspace}
\newcommand{\gencast}{GenCast\xspace}

\newcommand{\tenthdegree}{\ang{0.1}\xspace}
\newcommand{\fifthdegree}{\ang{0.2}\xspace}
\newcommand{\quarterdegree}{\ang{0.25}\xspace}
\newcommand{\erafivenativedegree}{\ang{0.28125}\xspace}
\newcommand{\onedegree}{\ang{1}\xspace}
\newcommand{\hresfczero}{HRES-fc0\xspace}

\newcommand{\cP}{\mathcal{P}}

\begin{abstract}
Machine learning (ML)-based weather models have rapidly risen to prominence due to their greater accuracy and speed than traditional forecasts based on numerical weather prediction (NWP), recently outperforming traditional ensembles in global probabilistic weather forecasting.   This paper presents \ourmodel, a simple, scalable and flexible modeling approach which significantly outperforms the current state-of-the-art models. \ourmodel generates ensembles via learned model-perturbations with an ensemble of appropriately constrained models. It is trained directly to minimize the continuous rank probability score (CRPS) of per-location forecasts. It produces state-of-the-art ensemble forecasts as measured by a range of deterministic and probabilistic metrics, makes skillful ensemble tropical cyclone track predictions, and captures joint spatial structure despite being trained only on marginals.

\end{abstract}

\begin{document}

\maketitle

\section{Introduction}

Accurate, reliable weather forecasts help protect and enrich human life, and provide tremendous value across the public and private sectors. In recent years, machine learning (ML)-based weather forecasting has made rapid advances, now rivaling and surpassing the top operational forecast models based on traditional, physics-based methods~\citep{keisler2022forecasting,kurth2022fourcastnet,bi2022pangu,lam2022graphcast,bi2023accurate,chen2023fengwu,li2023fuxi,han2024fengwughrlearningkilometerscalemediumrange,lam2023learning,zhong2024fuxi,bodnar2025}. These advances have been driven by increasingly powerful ML methods which are driving much of the wider artificial intelligence (AI) revolution, as well as high-quality datasets of historical weather~\citep{hersbach2020era5,ifs-manual-cy46r1-ens}, and invaluable benchmarks that allow for measuring progress and comparing techniques~\citep{ECMWF_IFS_49r1_Scorecard_2024,rasp2020weatherbench,rasp2023weatherbench}.

While most of the aforementioned ML weather models have been \textit{deterministic}, typically modeling something close to the ``average'' weather forecast, the applied weather community has increasingly emphasized their probabilistic (ensemble) models, such the European Centre for Medium-Range Weather Forecasts (ECWMF)'s ENS \citep{ifs-manual-cy46r1-ens} and NOAA's GEFS \citep{zhou2022development}, to capture not only the most likely future weather conditions, but the range of probable conditions that may unfold. The reason probabilistic forecasts are necessary is that the underlying physical dynamics are complex and non-linear, and our weather observations are only partial, so unobserved factors can have substantial influence on future weather conditions, and even the best weather models cannot make precise forecasts with perfect accuracy. 
Probabilistic weather models can inform critical decisions, by allowing decision-makers to plan for both the most likely scenario, as well as less probable scenarios, e.g. dangerous storms, which are important to anticipate and prepare for. 

Advances in ML-based probabilistic weather forecasting are more nascent, with early attempts leveraging a range of generative modeling techniques including generative adversarial networks~\citep{ravuri2021skilful}, flow-matching~\citep{couairon2024archesweathergen}, variational auto-encoders~\citep{zhong2024fuxi,oskarsson2024probabilistic}, as well as hybrid ML-physics models~\citep{kochkov2023neural,kochkov2024neural}. ML-based global probabilistic forecasts have only recently begun to compete with operational systems~\citep{kochkov2023neural,lang2024aifs}. The first model to clearly outperform the top physics-based operational system, ECMWF's ENS, was \gencast~\citep{price2023gencast,price2025probabilistic}. While \gencast is also much more computationally efficient than traditional NWP, it is relatively slow compared to some other ML-based models, and does not provide a straightforwardly adaptable framework for weather models to move beyond dense gridded data.

Here we present \ourmodel, a new combined architectural, training, and inference approach for probabilistic modeling of weather, which is faster, more flexible, and has greater performance than \gencast. We train \ourmodel on ECMWF's ERA5 reanalysis and operational HRES initial conditions to optimize the Continuous Ranked Probability Score (CRPS), a popular top-line metric for probabilistic weather forecasts, similar to several related approaches which include or modify CRPS in their objective, including NeuralGCM~\citep{kochkov2023neural}, Graph-EFM \citep{oskarsson2024probabilistic}, FuXi-ENS \citep{zhong2024fuxi}, and AIFS-CRPS~\citep{lang2024aifs}.
It models epistemic and aleatoric uncertainty with distinct mechanisms, model ensembles~\citep{lakshminarayanan2017simple} for the former, and a method which is related to other work on stochastic functions ~\citep{huang2017arbitrary,alet2024functional,lang2024aifs} for the latter.
Our results show \ourmodel comprehensively outperforms \gencast (and ENS) across a benchmark of deterministic and probabilistic metrics. Its forecast distributions generally have similar-to-better calibration than previous models, it predicts extremes as well or better, and it has better joint distribution structure in most cases we examined. \ourmodel's tropical cyclone track forecasts are also significantly better ($p < 0.05$), for both mean trajectories and track probabilities.

\begin{figure}[t]
    \centering\includegraphics[width=1.0\linewidth]{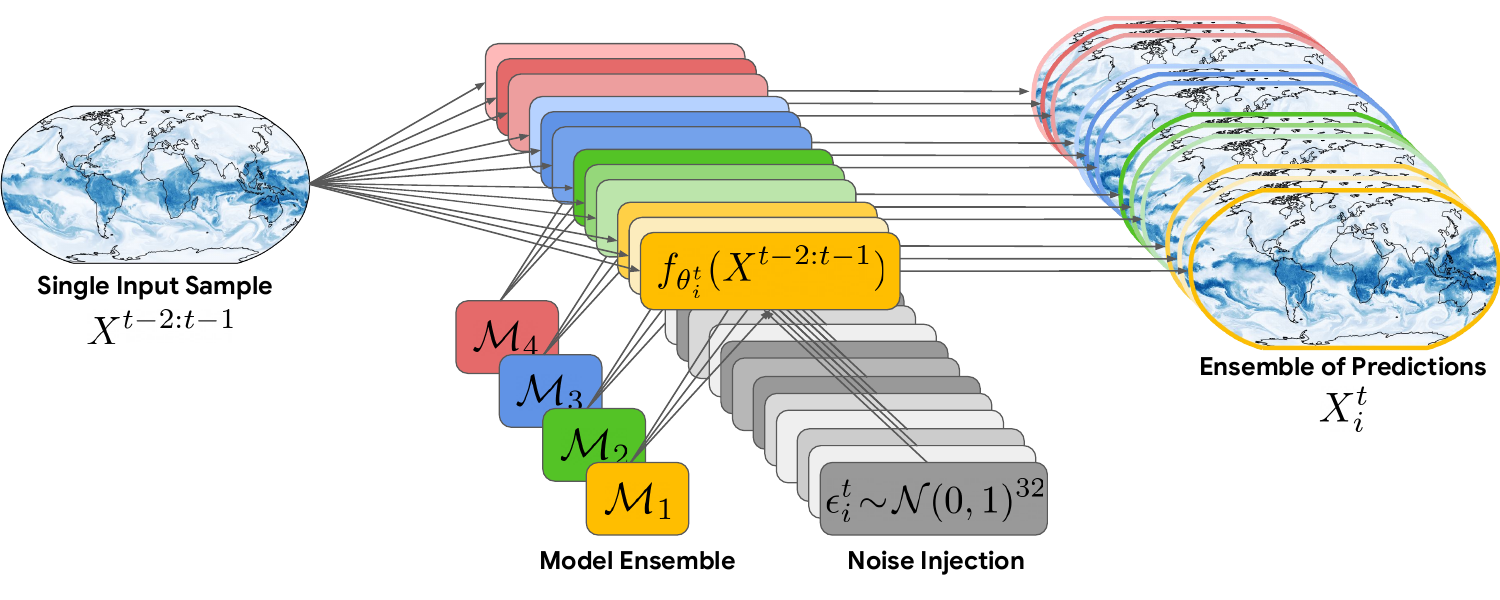}
    \caption{
    An overview of the \ourmodel generative process for producing a single step of a forecast ensemble from a single pair of input frames ($X^{t-2:t-1}$). Diversity is introduced at two levels, modeling aleatoric and epistemic uncertainty respectively. For a given model $\mathcal{M}_j$, aleatoric uncertainty is introduced at each step of a forecast trajectory by sampling a low-dimensional noise vector $\epsilon^t_i$ used for parameter-shared conditional normalization during the model forward pass. This can be interpreted as applying a perturbation to the neural network weights to obtain $\theta_i^t$, and hence, as sampling the parameters of the neural network. To generate $N$ ensemble members with aleatoric uncertainty, we simply condition on $N$ different $\epsilon_i^t$ independently. The epistemic uncertainty is modeled by ensembling outputs of multiple models $\mathcal{M}_j$ (each with their own $\{\theta^*_j, \Delta_j\}$ parameters) trained independently, each of which generates a subset of ensemble members according to the procedure described above. 
    }
    \label{fig:figure_1}
\end{figure}

\section{Methods}
\subsection{Problem formulation}~\label{subsec:problem_formulation}
The ensemble weather forecasting problem we tackle in this work is to draw samples from the joint probability distribution, $p(X^{1:T} | X^{\leq0})$, over $T$-step weather trajectories $X^{1:T}$ conditional on prior weather. In its fullest description, the ensemble forecasting problem requires conditioning directly on weather observations and measurements made during a time window that leads up to the initialization time $t=0$. However, following standard practice in operational weather forecasting, we assume for now that the conditioning on observations is handled as part of a data assimilation process that we take as a given, generating the initial states $X^{\leq0}$ on which our predictive distribution conditions.

When modeled as an $n$-th order Markov system, this distribution over whole trajectories factorizes into the product of distributions over each successive state, conditional on the prior $n$ states. In this work we choose $n=2$, and so we have
\begin{align}
p(X^{1:T} | X^0, X^{-1}) = \prod_{t=1}^T p(X^{t} | X^{t-2:t-1}).
\end{align}
We thus train a model to sample from $p(X^{t} | X^{t-2:t-1})$, 
and then for each $t \in 1{:}T$ we sample $X^t$ auto-regressively conditional on the previous two steps $X^{t-2:t-1}$. Repeating this process multiple times generates an ensemble of trajectories.

In the results that follow, each $X^t$ consists of 6 atmospheric variables at 13 vertical pressure levels and 6 surface variables (see \cref{tab:app:variables}), at each point on a \quarterdegree latitude-longitude grid, at timestep~$t$. We target 15 day forecasts with timesteps of 6 hours.

\subsection{Functional generative networks: skillful joint distributions through functional variability}~\label{sec:fgn}
Sources of uncertainty are commonly distinguished as being either epistemic or aleatoric. Epistemic uncertainty is the uncertainty about the model due to finite training data and an imperfect learning algorithm. Aleatoric uncertainty is due to the inherent irreducible randomness in a process from the perspective of a model. 
We propose \textit{functional generative networks}~(FGNs), a simple, scalable approach which models both the epistemic and aleatoric uncertainty in the distribution of forecasts, by expressing uncertainty through functional variability. FGNs learn to model the joint distribution over the outputs while training on a loss that applies only to the marginal distributions, in a suitably constrained setting.

\subsubsection{Modeling epistemic uncertainty with multi-model ensembling}

One can account for epistemic uncertainty by drawing samples from the posterior predictive distribution,
\begin{align}
p(X^{1:T} | X^{\leq 0}, \mathcal{D}) &= \int p(X^{1:T} | X^{\leq 0}, \mathcal{M}) \, p(\mathcal{M} | \mathcal{D})\mathrm{d}\mathcal{M}
\end{align}
which marginalizes over some posterior uncertainty about the model $\mathcal{M}$ conditional on our training dataset $\mathcal{D}$.
We adopt a common approximation known as deep ensembles \citep{lakshminarayanan2017simple}, in which we train multiple models $\mathcal{M}_1, \ldots, \mathcal{M}_J$ from independently-initialized parameters, and then later ensemble their predictions. Its connection to Bayesian inference has been studied elsewhere~\citep{wilson2020bayesian,wild2023rigorous}, but here we apply it informally to approximate Bayesian inference.

In this work we train $J=4$ models. To ensemble their predictions, we generate an equal number of ensemble member trajectories from each model, using the same model for all timesteps of a trajectory.

\subsubsection{Modeling aleatoric uncertainty with learned functional perturbations}

The weather states we model from ECMWF's ERA5 and HRES analysis do not fully capture the state of the weather system. It is common to use stochastic dynamics to model uncertainty due to unresolved phenomena, and from the perspective of the model this can be viewed as aleatoric uncertainty.

We incorporate this aleatoric uncertainty in our model by sampling functions.
Concretely, we model different forecast samples $X_i^t$ of $p\left(X^t|X^{t-2:t-1}\right)$ as coming from neural networks $f_{\theta_i^t}$ whose parameters $\theta_i^t$ are sampled from a learned distribution: $X_i^t=f_{\theta_i^t}(X^{t-2:t-1})$, where $\theta_i^t\sim\cP_\mathcal{M}(\theta)$
independently for each ensemble member $i$ and timestep $t$.

\citet{alet2022learning} justifies why modeling aleatoric uncertainty in parameter space can lead to more structured variations than modeling it in input or output space. This structure is due to the reuse of sampled parameters across spatial (or other relevant) dimensions in architectures with parameter-sharing layers. Similar insights have been leveraged in RL exploration~\citep{fortunato2017noisy}, 
generative adversarial networks~\citep{saatci2017bayesian}, and test-time adaptation~\citep{sun2020test,bibas2019deep,wang2020tent,alet2021tailoring}.

This idea is analogous to stochastic physics, where parameters inside a numerical weather prediction system are perturbed to obtain an ensemble, in addition to initial condition sampling \citep{ollinaho2017towards}. ML weather models have naturally explored perturbing models in order to make them probabilistic~\citep{kochkov2023neural,lang2024aifs}, although these have generally used spatially-independent model perturbations.

\subsubsection{Reparameterization trick and CRPS marginal training}~\label{sec:crps}
To learn the distribution $\cP_\mathcal{M}(\theta)$, we use the well-known reparameterization trick~(see~\citealt{kingma2013auto,pmlr-v32-rezende14,titsias2014doubly}, as well as \citealt{mohamed2020monte} for an instructive introduction): $\theta=\theta^*+\Delta\cdot\epsilon$, where $\Delta$ is a potentially-structured matrix used to map Gaussian noise $\epsilon$ to a perturbation in $\theta$ space. $\mathcal{M} := \{\theta^*, \Delta\}$ thus consists of $\theta^*$, acting as the mean $\theta$ and describing a deterministic function, and a matrix $\Delta$ controlling the covariance. At inference time, we make the modeling assumption that we can compose both types of uncertainty by sampling one of multiple $\mathcal{M}_j$ trained independently, and then sampling $\epsilon^t_i$ to get  $\theta^t_i = \theta^*_j + \Delta_j \cdot \epsilon_i^t$, as shown in Figure~\ref{fig:figure_1}.

We use the Continuous Ranked Probability Score (CRPS) as our training objective, which is  a strictly proper scoring rule for univariate distributions \citep{gneiting2007strictly}. For a cumulative density function (CDF) $F$ and target $y\in\mathbb{R}$,
\begin{align}\label{eq:cdf_crps}
    \text{CRPS}(F,y) = \int_{-\infty}^\infty \left(F(z)-\large{\mathbb{1}}[z\geq y]\right)^2 dz .
\end{align}
For an empirical distribution coming from samples $x^{1:N}$ this is equal to: $\frac1N\sum_n|x^n-y|-\frac{1}{2N^2}\sum_{n,n'}|x^n-x^{n'}|$.
However, this is a biased estimator of the CRPS of the underlying distribution $F$. For an unbiased CRPS estimate we can use the `fair' CRPS estimator \citep{zamo2018estimation},
\begin{align}\label{eq:crps}
    \text{fCRPS}(x^{1:N},y) &:= \frac1N\sum_n|x^n-y|-\frac{1}{2N(N-1)}\sum_{n,n'}|x^n-x^{n'}|.
\end{align}
Whenever we refer to CRPS we mean this fair estimator, unless stated otherwise (e.g. in Section~\ref{sec:eval}).

In practice during training we use $N=2$, and average over all locations, variables, and levels,
\begin{align}\label{eq:loss}
    \mathcal{L} &:= \frac1{|D|} \sum_{d} \frac1{G} \sum_i a_i \text{fCRPS}(x^{1:2}_{i,d}, y_{i,d})
\end{align}
where $d$ indexes into the dataset batch $D$, $i$ indexes into the $G$ variable-level-location tuples and $a_i$ is the corresponding loss weighting which we take from \citet{lam2022graphcast} and \citet{price2023gencast} (with the exception of the geopotential variables, for which we halved the loss weight to help tame overfitting as we scaled the model size).

As noted in the introduction, a number of other works have also included or modified the CRPS objective when training ML-based weather models \citep{kochkov2023neural,oskarsson2024probabilistic, zhong2024fuxi,lang2024aifs}.  Our approach is also closely related to work on Scoring Rule minimization \citep{pacchiardi2024probabilistic}, DISCO Nets~\citep{bouchacourt2016disco} and MMD nets~\citep{gritsenko2020spectral}, where ensemble models are also trained with CRPS from noisy inputs. However, these latter works typically do not yield state-of-the-art joint distributions in high-dimensional problems. In this work, we show that modeling the noise in parameter space rather than the original inputs or outputs can be a simple, effective method to learn skillful joint distributions in a very high dimensional setting, given a suitably strong algorithmic inductive bias.

\subsection{Model Architecture}

Each of the constituent models in the \ourmodel model-ensemble is implemented with a very similar neural network architecture to that of the denoiser in \gencast~\citep{price2025probabilistic}, with a GNN encoder/decoder mapping from the lat/lon grid to a latent space defined on a spherical 6-times-refined icosahedral mesh, and a graph-transformer processor which operates on the nodes of this mesh. 

There are important differences, however, between the \ourmodel and \gencast architectures. \ourmodel is a larger model, with $\sim$180m parameters per model seed, compared to \gencast's $\sim$57m in total. \ourmodel has latent dimension 768 and 24 layers in its graph-transformer processor, compared with latent dimension of 512 and 16 layers in \gencast, and produces forecasts with a 6-hour timestep, whereas \gencast has a 12-hour timestep\footnote{A number of other very minor differences in the encoder architecture are outlined in Appendix \ref{sec:app:architecture_changes}.}. In Appendix \ref{sec:ablations}, we ablate the impact of these model ensembling and model size differences.

One key modeling choice was to use conditional normalization---which in \gencast are used to condition on the corresponding diffusion noise level $\sigma$---as the means of perturbing the model. To be precise, in \ourmodel a global noise vector $z \sim \mathcal{N}(0,1)^{32}$ is passed into all of the network's conditional layer-norm layers, such that sampling different vectors $z$ for each ensemble member is what generates the variance across the ensemble.

This approach has two important features: 1) a low-dimensional noise source, which 2) is globally applied across all layers, and with learned conditional normalization parameters which are shared across the spatial dimensions of the model (i.e. mesh and grid node dimensions). Together these constrain the output distribution in a way that encourages the model to generate globally coherent variability despite being trained on a marginals-only loss. This is discussed in more detail in Section~\ref{sec:discussion}.  

We also note that conditional normalization has had success in other contexts for semantically-coherent functional modulation, for example in diffusion to condition on noise levels, as noted above, or in generative models of images and audio to condition on style or voice \citep{chen2021adaspeech, karras2019style}. \citet{lang2024aifs} also use conditional normalization to inject stochasticity into their weather model, but do so with location specific conditioning, which adds a high spatial frequency component to the noise injection. 

\subsection{Auto-regressive training}
\ourmodel is initially trained with a `single-step' loss only. The final stages of training, however, involve autoregressive (AR) rollouts as part of each training step, where the model produces a number of forecast steps autoregressively, and the loss is averaged over all rollout steps, with gradients propagated back through the rollout. \ourmodel is finetuned on rollouts up to 8 steps. The exact training stages and their hyperparameters are described in Appendix \ref{sec:schedule}. We find AR training to be helpful, but not essential, to performance---skillful joint forecast distributions are also achieved without it (see Appendix \ref{sec:ablations}).

\section{Experiments}
\subsection{Training and evaluation data}
We make use of two datasets: ERA5 and \hresfczero to train \ourmodel (details in Appendix \ref{sec:data}). We use ERA5 for ``pre-training'', because it is a large corpus of historical weather reanalysis, from a single NWP that was used operationally in 2016, but is not available in real-time for operational settings. We use \hresfczero for ``fine-tuning'', because it is a smaller dataset that is comprised of analysis from NWPs used operationally from 2016 to the present, and is available in real-time in operational settings. While developing our model, we evaluated against \hresfczero as ground truth, using 2022 as the validation year. Once finished, we froze the architecture and training protocol and trained a model on data up to and including 2022. The resulting model was evaluated on \hresfczero from 2023 as the test year, yielding the results reported below. 

\subsection{Baseline: \gencast}~\label{subsubsection:gencast}
We use \gencast~\citep{price2023gencast, price2025probabilistic} as a strong baseline model, against which we compare \ourmodel's performance. \gencast is the first probabilistic machine learning based weather prediction (MLWP) model to significantly outperform the top NWP system, ECMWF's ENS, at high resolution, and
represents the state-of-the-art for probabilistic medium-range weather forecasting. 
\gencast is a diffusion model, and produces a single forecast trajectory by sampling one forecast timestep at a time (using an iterative denoising process), and rolling out step by step by feeding each output step back in as input to generate the subsequent step. It can generate an ensemble of stochastic trajectories by sampling single trajectories independently. It is trained explicitly to model the joint distribution over the weather at the next timestep, given the weather at the previous two timesteps. \gencast operates at \quarterdegree resolution, and generates ensembles of 15 day prediction with 12-hour timesteps.

While \gencast was trained and evaluated on ERA5 reanalysis data in \citet{price2025probabilistic}, it has since been finetuned on operationally available \hresfczero data, as part of Google's WeatherNext family of operational models where it is called WeatherNext Gen. Here we focus our evaluations on the operational setting, and thus compare performance to the WeatherNext Gen forecasts, though we refer to the model as \gencast throughout this paper\footnote{\citet{price2025probabilistic} evaluated \gencast against ERA5 as ground truth, while ENS was evaluated against \hresfczero. Appendix \ref{app:ens} compares the WeatherNext Gen forecasts which we use as the baseline in this paper, against ENS for the full year of 2023, evaluating both models against \hresfczero. The figures show that the results presented in \citet{price2025probabilistic} carry over to the operational setting.}. These forecasts are publicly available via \href{https://developers.google.com/earth-engine/datasets/catalog/projects_gcp-public-data-weathernext_assets_126478713_1_0}{Earth Engine} and  \href{https://console.cloud.google.com/bigquery/analytics-hub/discovery/projects/gcp-public-data-weathernext/locations/us/dataExchanges/weathernext_19397e1bcb7/listings/weathernext_gen_forecasts_193b7e476d7?inv=1&invt=AbzlsA}{BigQuery}\footnote{Note that another small difference between WeatherNext Gen and the one in \citet{price2025probabilistic} is that this version also predicts 100m wind. However, this has minimal effect on its skill on other variables and thus does not matter for the comparisons in this paper.} where each year of data has been generated with a model finetuned on HRES-fc0 from 2016 up to the start of that year. As such, we make use of the 2022 and 2023 forecasts for validation and test years respectively.

\subsection{Training and inference efficiency}

Training \ourmodel takes approximately 3 wall clock days, using a combined total of 490 TPUv5p and TPUv6e days of compute, for each of the four models in the model-ensemble (see Appendix \ref{sec:schedule} for details).

For inference, generating a single 15-day forecast trajectory takes just under 1 minute on a single TPU v5p, and every ensemble member can be generated in parallel. We highlight that despite being a substantially larger model than \gencast, and generating twice as many forecast frames due to its 6-hour timestep, generating a 15-day forecast is still 8 times faster. This is due to the fact that \ourmodel only requires a single forward pass through the network to generate a forecast, compared with the multiple forward passes needed for the iterative refinement process of a diffusion model.

\section{Evaluation Methodology}
To evaluate \ourmodel, we follow a similar methodology as that used to evaluate \gencast in \citet{price2025probabilistic}. One key difference is that in this paper all models are initialized with and evaluated against \hresfczero, at all four initialization times of the day (00z, 06z, 12z, 18z), quantifying the performance of these ML-based weather models in realistic operational settings. We also evaluate ensembles of size 56 for both models (in the case of \ourmodel, each model seed generates 14 members).
We refer the reader to the appendix of \citet{price2025probabilistic} for a detailed description and motivation of the various metrics and their computation, including statistical significance testing.

\subsection{Evaluating marginal forecast distribution}\label{sec:eval}

We first evaluate the predictive skill of marginal forecast distributions.
Marginal distributions are pixel-wise, single-variable distributions most relevant to users interested in the weather at a given location and time.
We assess these marginal distributions using the RMSE of the ensemble mean, and the CRPS of the ensemble, computed at each location and then globally averaged, for each variable and pressure level. We use the biased versions of these metrics because the \ourmodel model ensemble violates some underlying assumptions of their bias-corrected variants\footnote{Specifically, when we enforce that \ourmodel generates an equal number of forecasts per model seed, this violates the independence assumption underlying the unbiased estimators.}.

Additionally, we assess the reliability and spread of these marginal distributions using spread-skill ratios, where well-calibrated forecasts will exhibit spread-skill ratios close to 1 \citep{leutbecher2009diagnosis,fortin2014should}.
Subject to some mild, standard assumptions, spread-skill ratios less than 1 indicate under-dispersion (over-confidence) on average.
Conversely, spread-skill ratios greater than 1 indicate over-dispersion (under-confidence) on average. 

Finally, we quantify the value of a probabilistic model for decision-making relative to extreme weather scenarios using the Relative Economic Value (REV) \citep{Richardson_2006,richardson2000skill,wilks2001skill}.
We use REV to quantify the benefit of using forecast distributions from a given model to decide whether to prepare (or not) for the event that a variable will exceed a given threshold. We evaluate REV for different C/L ratios where C is the cost of preparing, and L is the loss incurred when encountering the event without preparation (but which could have been avoided if one had prepared). We compute REV curves for various thresholds corresponding to high and low percentiles of the climatological distribution (computed for each variable on a per-latitude, longitude, month-of-year and time-of-day basis, using 24 years of ERA5 from 1993 to 2016). REV is normalized relative to the value of a climatological forecast (REV=0) and a perfect forecast (REV=1).

\subsection{Evaluating correlation structure of the joint distribution}

Weather forecasts are used in many applications where the correlation between various parts of the predictions are important. For instance, one might be interested in the total amount of precipitation accumulated in an entire region, or in how several variables can be used to compute another quantity of interest.
We assess the quality of the correlation structure of the joint forecast distributions via a number of standard proxies, each of which quantifies specific aspects of this joint distribution.

First, we assess the spatial correlations and dependencies in the forecast distribution by performing probabilistic evaluation on spatially pooled predicted fields.
We follow the pooling protocol described in `Spatially pooled CRPS evaluation' in Methods in~\citet{price2025probabilistic}, evaluating max-pooled and average-pooled CRPS for pool sizes ranging from \SI{120}{km} to \SI{3828}{km}.
Note that when spatial dependencies exist,  minimizing the CRPS of the marginal predictions is not sufficient to minimize the CRPS of the pooled predictions. Thus, the pooled CRPS provides a useful metric to assess the spatial structure of forecasts' joint distributions.

Second, we evaluate the cross-variable dependencies in the forecasts by computing the CRPS for two derived quantities of interest: 10m wind speed (derived from the $\text{10u}$ and $\text{10v}$ predicted fields), as well as the difference in geopotential at pressure levels \text{300}hPa and \text{500}hPa, which is a critical quantity for tropical cyclones predictions as well as weather more generally. Skill in forecasting these derived quantities relies on correctly modeling the dependencies of their constituent predicted fields. 

Finally, we report the spherical harmonic power spectra of the forecasts and compare it to those of the ground truth \hresfczero. Power spectra are a standard, albeit imperfect, measure of physical plausibility. They are an important tool to diagnose blurring, a common but ``unphysical'' way deterministic ML-based weather models achieve lower RMSE.

\subsection{Evaluating tropical cyclone tracks}

We evaluate the predictive skill of the models at the task of cyclone track prediction using two metrics: the position error of the ensemble mean trajectories, as well as the REV of forecast track probabilities, using the protocol from \citet{price2025probabilistic}.

Position error is calculated relative to named cyclones from the International Best Track Archive for Climate Stewardship (IBTrACS, \citealt{knapp2010ibtracs}) which we treat as ground truth.
Forecast cyclone tracks are produced by applying the Tempest Extremes tracker \citep{ullrich2021tempestextremes} to the forecasts of each model, using hyper-parameters detailed in Methods and Supplementary Information of \citet{price2025probabilistic}.
Forecast tracks are then paired with ground truth tracks if their initial position is within \SI{100}{km} at lead time zero.
We consider the mean position prediction to ``exist'' while 50\% or more ensemble members continue to forecast a cyclone (as measured by the continued existence of a Tempest Extremes track for that member).
We then average the position error of ensemble mean cyclone positions over the intersection of named cyclone and lead time pairs for which all models provide a track, as is standard practice. 

Position error is a `deterministic' evaluation in the sense that it does not evaluate any aspect of the forecast distribution beyond its mean. 
For a probabilistic evaluation, we report the REV of track probability predictions i.e., the predicted probability that a cyclone center will pass through a given place at a given time. Specifically, we compute \ang{1}~resolution track probability heatmaps, and the predicted probability in each \ang{1}~cell of the heatmap is the fraction of ensemble members forecasting a cyclone center within that cell.

\section{Results}
\subsection{Results on marginal predictive skill and calibration}

\begin{figure}[p]
    \centering
    \includegraphics[width=\textwidth]{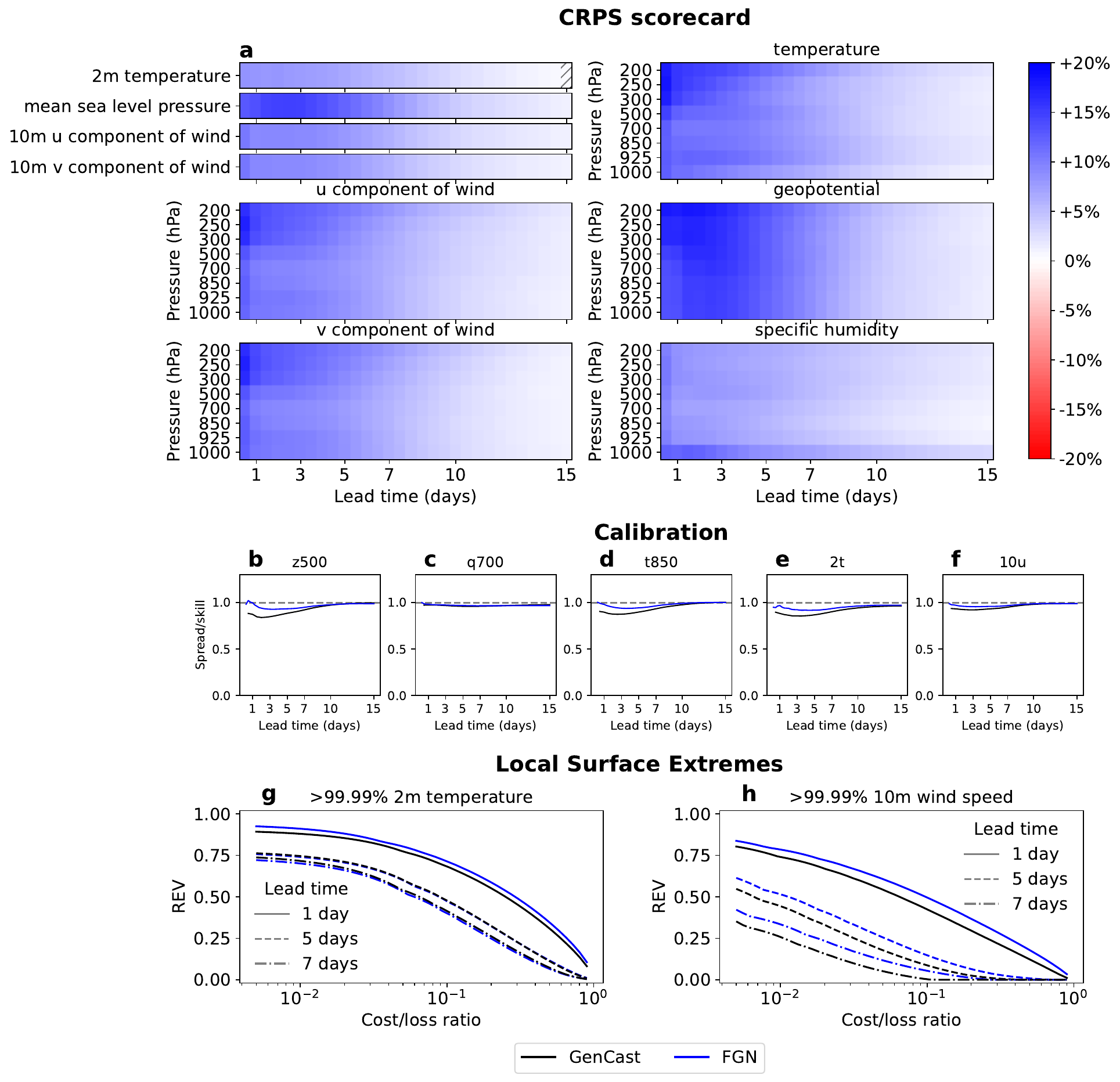}
    \caption{\textbf{\ourmodel produces more skillful marginal forecasts than \gencast while preserving good calibration}. (a) A scorecard comparing CRPS achieved by \ourmodel and \gencast, where blue cells indicate \ourmodel outperforms \gencast, red denotes where \gencast is better, and hatched regions indicate where the difference in performance was not statistically significant. \ourmodel is significantly better  ($p < 0.05$) in 99.9\% of cases, with an average improvement of 6.5\%. (b-f) Spread skill plots for a number of variables across all lead times. \ourmodel maintains a spread skill ratio very close to 1 across all lead times, indicating well calibrated ensemble spread. (g, h) REV for forecasts of extreme high (>99.99th percentile) 2m temperature and 10 wind speed, at lead times of 1, 5, and 7 days. \ourmodel achieves better REV than \gencast in the case of 10m wind speed, and matches performance on 2m temperature.}
    \label{fig:marginals}
\end{figure}

Figure~\ref{fig:marginals}a shows that \ourmodel achieves better (lower) CRPS than \gencast on 99.9\% of the targets, with improvements of up to 18\%, and an average of 6.5\%. The results are shown in the form of a scorecard, where blue cells indicate those variable-level-leadtime combinations where \ourmodel achieved better CRPS, while red would indicate that \gencast performed better. Hatched regions indicate where the difference in performance was not statistically significant ($p \geq 0.05$). We see similar results on  ensemble mean RMSE, with \ourmodel outperforming \gencast in 100\% of variable-level-leadtime combinations, with improvements of up to 18\% and an average of 5.8\%, as shown in the scorecard in \cref{fig:sup:rmse} in Appendix~\ref{sup:extended results}. These results indicate \ourmodel has greater forecast skill \gencast in its marginal forecast distributions\footnote{Note, due to our lack of confidence in the quality of ground-truth precipitation data, we exclude precipitation results from
our main figures, and refer readers to Figures~\ref{fig:sup:precip_12} and \ref{fig:sup:precip_24}.}.

\cref{fig:marginals}b-f shows that \ourmodel is also generally very well calibrated, with spread-skill scores close to 1 at all lead times. \gencast exhibits good but slightly poorer calibration over the first 5 - 7 days lead time, relative to \ourmodel\footnote{We note that the calibration of this operational \gencast model, while still good overall, is slightly worse than that of the ERA5 version of the model presented in \citet{price2025probabilistic}.}.

\ourmodel also exhibits as good or better REV for predictions of extreme weather. Figure~\ref{fig:marginals} shows REV for predictions of 2m temperature (\ref{fig:marginals}g) and 10m wind speed (\ref{fig:marginals}h) exceeding the 99.99th percentile, with \ourmodel matching \gencast on the former, and outperforming \gencast on the latter. Results on more variables and thresholds are shown in 
\cref{fig:sup:rev_high_wind,fig:sup:rev_high_2t,fig:sup:rev_low_2t,fig:sup:rev_low_msl}.

\subsection{Results on structure of the joint distribution}

\begin{figure}[p]
    \centering
    \includegraphics[width=\textwidth]{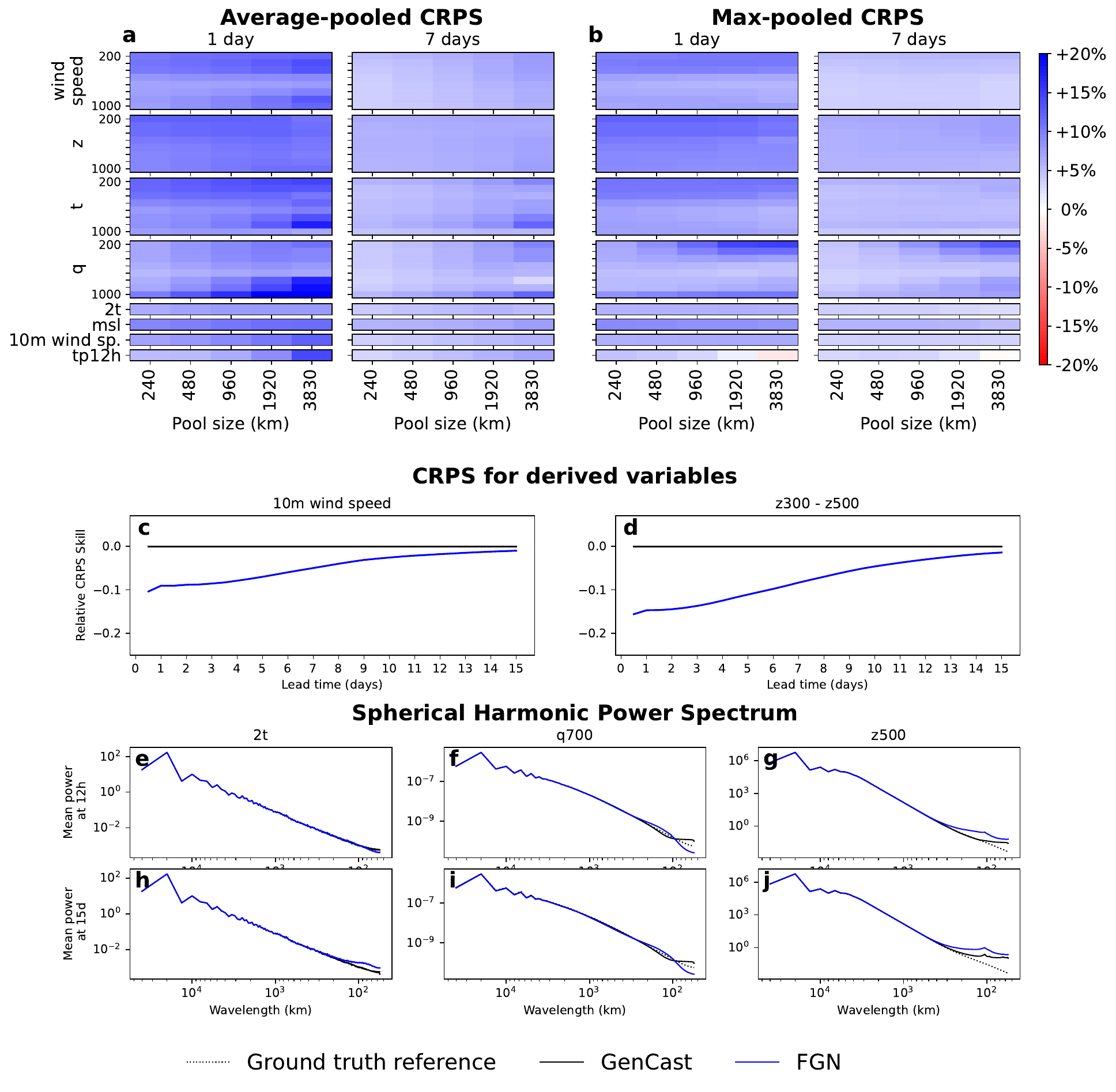}
    \caption{\textbf{\ourmodel produces skillful joint forecast distributions}. 
    (a, b) CRPS scorecards evaluating average-pooled (respectively  max-pooled) fields, for different pool sizes, at 1 and 7 day lead times. \ourmodel achieves better CRPS in average-pooled evaluations in 99.9\%  (respectively 99\%) of all pool-size, variable, leadtime, level combinations, evidencing that \ourmodel skillfully models spatial correlations in the forecast distribution.
    (c, d) Evaluations of skill on two quantities of interest derived from predicted fields. \ourmodel{} obtains $10.4\%$ and $15.6\%$ better CRPS on 10m wind speed and $z300 - z500$ at short lead times respectively, with this improvement decreasing as lead time increases. This shows \ourmodel is able to capture across-variable dependencies despite not being directly trained on them.
    (e - j) Spectra of \ourmodel{} compared to \gencast, and the ground truth. In some variables (e.g. z500), \ourmodel{} has a spike at the mesh frequency and some additional high frequency content, more so than \gencast, but orders of magnitude smaller than the dominant powers in the signal.
}
    \label{fig:joints}
\end{figure}

\ourmodel achieves average improvements of 8.7\% and 7.5\% over \gencast on average-pooled CRPS and max-pooled CRPS respectively across different spatial scales. Scorecards for 1 and 7 day lead times are shown in \cref{fig:joints}a and b. The observed general trend of these results are maintained across other lead times, with the average improvement decreasing as lead time increases towards 15 days. In total, across all level, variable, and pool-size combinations across all lead times, \ourmodel achieves better average-pooled and max-pooled CRPS in $99.9\%$ and $99\%$ of cases respectively. These results show that \ourmodel succeeds in capturing certain spatial dependencies in the joint distribution.

\cref{fig:joints}c and d shows that \ourmodel is also on average $4.8\%$ more skillful than \gencast when evaluating 10m wind speed predictions derived from its raw `10u' and `10v' predictions, and $7.8\%$ better on forecasts of z300 $-$ z500, derived from its z300 and z500 predicted fields. Improvements range from $10.4\%$ and $15.6\%$ respectively at short lead times, to $1-2\%$ at longer lead times, highlighting that \ourmodel succeeds in modeling these inter-variable dependencies. 

From the spherical harmonic power spectra shown in Figure~\ref{fig:joints}, \ourmodel's predictions generally match well with the ground truth. In particular, \ourmodel avoids the loss of power at high frequencies characterizing blurry predictions. This holds out to 15 day lead times, as shown in Figures \ref{fig:joints}e,h,f,i which plot the spectra of 2m temperature and specific humidity at 700hPa. In certain fields, in particular smoother fields like geopotential at 500hPa (Figures~\ref{fig:joints}g,j), \ourmodel's forecasts contain some additional high frequency signal and in certain cases minor artifacts around a wavelength of 100km, consistent with the artifacts explained in Section~\ref{sec:discussion}.
Similar phenomena were observed in a few variables in \gc\citep{lam2022graphcast}, but are more common with \ourmodel. There is also some minor build up in high frequency signal observed in other variables, like 2m temperature, at longer lead times, as is visible in Figure~\ref{fig:joints}h. Spectrum plots for additional variables at more lead times are shown in Figure \ref{fig:sup:spectra1} and \ref{fig:sup:spectra2} in Appendix \ref{sup:extended results}.

\subsection{Results on tropical cyclone tracks}

\begin{figure}
    \centering
    \includegraphics[width=\textwidth]{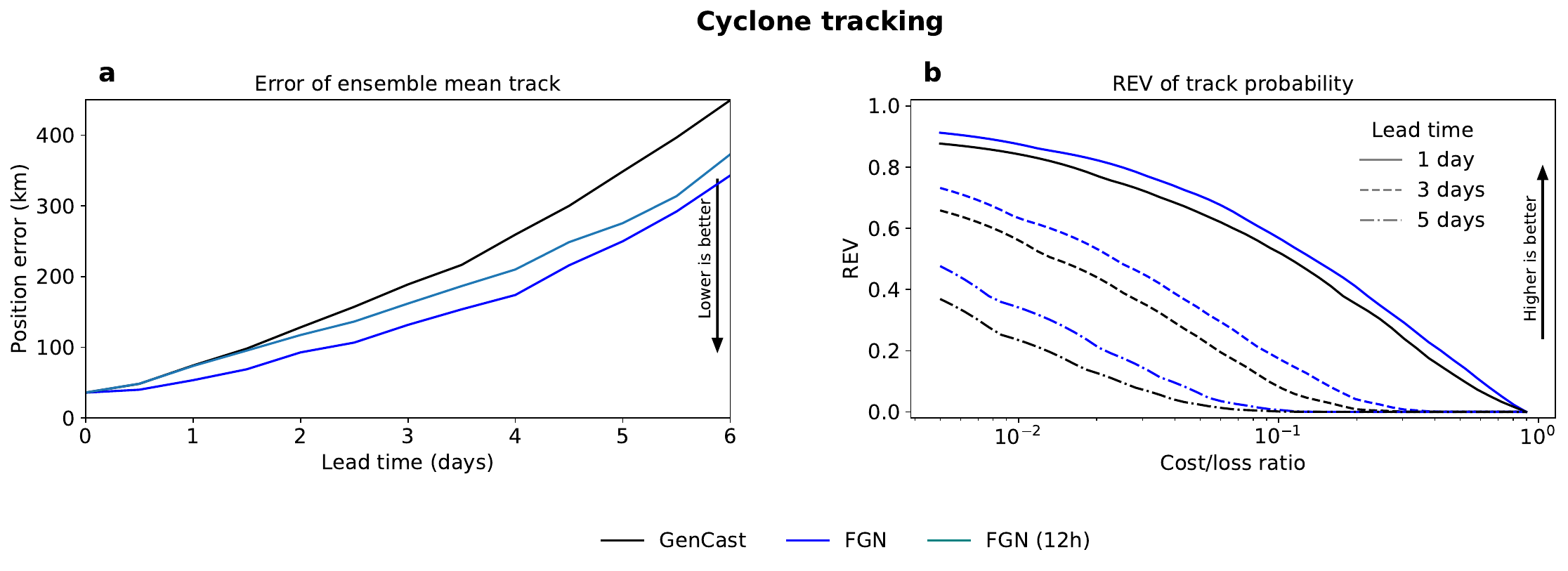}
    \caption{\textbf{\ourmodel achieves state-of-the-art cyclone track prediction.} (a) Position error of the ensemble mean track. \ourmodel achieves up to a 24h improvement in position error over \gencast. Some of this improvement is due to the TempestExtremes cyclone tracker working better on 6-hour timesteps than 12-hour timesteps. However, as shown in the plot, a 12h-step version of \ourmodel still achieves better position error beyond 2 day lead times. (b) REV of track probability predictions. \ourmodel exhibits better REV than \gencast across lead times of up to 5 days, across all cost/loss ratios at which either model is better than climatology.}
    \label{fig:cyclones}
\end{figure}

The ensemble mean track of \ourmodel{} is significantly more accurate than \gencast{}'s ($p < 0.05$), achieving on average approximately a 24 hour advantage in track accuracy from 3 - 5.5 day lead times. In other words, \gencast's accuracy at 2 days ahead is roughly achieved by \ourmodel 3 days ahead (Figure~\ref{fig:cyclones}a). Part of this improvement may be due \ourmodel predicting at 6-hour timesteps, which results in fewer `tracker jumps' and spurious tracker errors when applying the Tempest Extremes tracks to the forecasts to estimate the tracks. However, most of the improvement is due to \ourmodel's overall greater forecast accuracy. To establish this, we include a 12-hour step version of \ourmodel in Figure \ref{fig:cyclones}a to ablate the impact of the model timestep. This 12-hour step \ourmodel shows higher average error relative to the 6-hour step \ourmodel, but still lower error than \gencast from 2-day lead times onwards, with the gap increasing as lead time increases.

\ourmodel{} also produces better track probability forecasts than \gencast{}, exhibiting better REV (\cref{fig:cyclones}b) across almost all cost/loss ratios at which either model is better than climatology. The REV improvement is especially pronounced for smaller cost/loss ratios and longer lead times, which are most relevant for cyclone preparation decision making. This suggests that \ourmodel{}'s cyclone forecasts could offer significant additional value in operational cyclone-related decision-making settings.

\section{Discussion}\label{sec:discussion}
Our results show \ourmodel offers substantial improvements over previous ML-based probabilistic weather models, and sets a new state-of-the-art in ensemble forecasting. It outperforms \gencast with both the skill of its marginal forecasts, including on extreme weather, as well as its skill in forecasting quantities dependent on the dependency structure of the joint forecast distribution, notably including tropical cyclone tracks. It uses an ensemble of independently trained models to capture epistemic uncertainty, each of which captures aleatoric uncertainty through learned functional variability, which can be understood as perturbations applied to each models parameters when generating a forecast. It generates forecasts more efficiently than \gencast, and can easily accommodate auto-regressive training, which is a powerful method for optimizing models over longer lead times.

It is straightforward to extend \ourmodel to broader tasks than modeling dense, gridded analysis data. As a preliminary demonstration, we recently built an experimental extension of \ourmodel, which is trained to directly predict cyclones. Our internal evaluations indicate its performance is substantially better on predicting cyclone tracks, intensities, and sizes than that reported in the results above, which used the TempestExtremes tracker to extract tracks from \ourmodel forecasts. Its tropical cyclone forecasts are available on our website (\url{https://deepmind.google.com/science/weatherlab}).

Perhaps the most fascinating feature of our results is how well \ourmodel captures the covariance of the joint spatial distribution of the forecasts. 
Optimizing CRPS of the forecast marginal distributions, alone, does not guarantee good joint distributions~\citep{pic2025proper}. This is trivially shown---a collection of independent models, one model per location, predicting the weather at each location independently, cannot capture spatial correlations, even if each model minimizes CRPS on its own forecast targets. Nonetheless, as we show in Appendix \ref{sec:ablations}, even the most simple, stripped down version of \ourmodel---which is smaller (512 latent dimension and 16 layers), has only a single model seed, and no autoregressive training meaning a truly marginals-only training objective---still outperforms the prior state-of-the-art almost across the board.

The reason \ourmodel captures joint structure is likely a result of how stochasticity is injected and transformed by the model's architecture.
The source of all stochasticity in the full 87 million-dimensional 
output distribution is a single 32-dimensional vector passed into the network's conditional normalization layers. This means that for a given input, the function output has only 32 degrees of freedom, and the impact of traversing these degrees of freedom on the network's computations is shared across its mesh and spatial dimensions. 
Notice also that this low effective dimensionality is independent of the resolution of the output grid, or indeed whether the output is a fixed and regular grid at all---the same will apply across any combination of dense and/or sparse predictions. While this does not provide guarantees for getting good joint forecast distributions by optimizing only the CRPS of its marginals, it does provide some intuition. We speculate that under such heavy distributional constraints and with the inductive biases of \ourmodel's architecture, the easiest way for the model to jointly optimize the CRPS of all marginals is to try to model their inter-dependencies as well.

\begin{figure}[htb]
    \centering\includegraphics[width=0.5\linewidth]{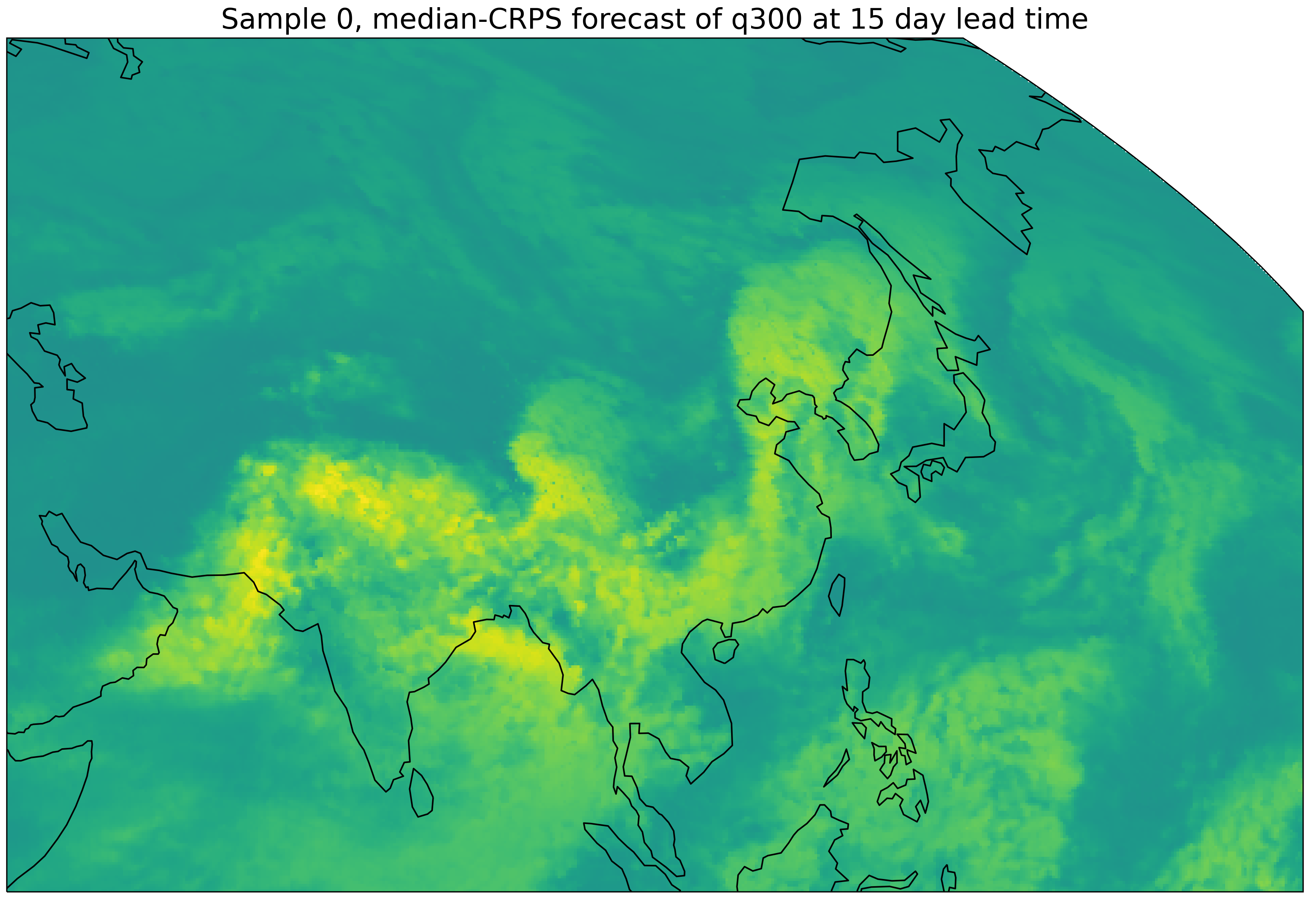}
    \caption{Visualization of a q300 forecast at 15 day lead time, zoomed in to highlight the `honeycomb' artifacts visible in this variable.}   \label{fig:artifacts}
\end{figure}

We also observed several weaknesses, which future work should address. One was that there can be subtle artifacts in the forecast states, which appear as visible `honeycomb' patterns corresponding to the processor's mesh structure. These artifacts tend to appear in higher frequency variables which were allocated smaller loss weights, specifically low pressure levels of specific humidity (see Figure~\ref{fig:artifacts}), though these appear as only very minor artifacts in the spectra of these fields. While forecasts of some smoother variables, like z500, exhibit more notable artifacts or `kinks' in their spectra, these do not manifest as visual artifacts at all, because high frequencies at which these `kinks' are present, contribute so little to the overall signal. The emergence of these artifacts may in fact be due to a slight \textit{over-constraining} of the output covariance structure which could enforce spurious local correlations. This would suggest that perhaps a slightly higher-dimensional noise input (and therefore higher-dimensional output manifold) might enable the forecasts to avoid these artifacts while still encouraging the learning of good joint structure. 

Another minor issue we observed was the potential for poor model seeds, which can result in some unstable forecast rollouts. Specifically, at one point during model validation, we found that a particular training seed produced a number of unstable rollouts, which was detected by examining the averaged spectra of the validation year (2022) forecasts. We removed this seed and retrained that particular model with a different seed, confirming that this model behaved as expected again using our validation set. This highlights the importance of careful evaluation of trained models before deployment.

Overall, our work shows that \ourmodel provides a general-purpose approach for modeling the different sources of uncertainty in weather forecasts, and capturing the joint spatial structure of forecast distributions. The \ourmodel framework is easy to work with and extend, and represents a key step forward in applying powerful statistical learning methods to one of the most complex and important learned simulation problems.

\FloatBarrier

\section{Acknowledgments}
In alphabetical order, we thank Aaron Abood, Keith Anderson, Boris Babenko, Luke Barrington, Maria Bauza, Juanita Bawagan, Molly Beck, Aaron Bell, Massimiliano Ciaramita, Marc Deisenroth, Ben Gaiarin, Zoubin Ghahramani, Olivia Graham, Raia Hadsell, Stephan Hoyer, Anna Koivuniemi, Paul Komarek, Petros Koumoutsakos, Elinor Kruse, Ian Langmore, Amy Li, Ken Mackay, Alberto Magni, Sunny Mak, Samier Merchant, Shakir Mohamed, Kevin Murphy, Bryan Norcross, Alexander Pritzel, Drew Purves, Anton Raichuk, Stephan Rasp, Jack Searby, Armin Senoner, Guy Shalev, Rachel Stigler, Alok Talekar, Gregory Thornton, Thomas Turnbull, Akib Uddin, Suhani Vora, Brian White, William White, Natalie Williams, and Fred Zyda for their advice, support and/or feedback on our work. We also thank ECMWF and NOAA for providing invaluable datasets to the research community, and experts at NHC, CIRA, UKMet, University of Tokyo, and Japan's Weathernews Inc. for their insights on tropical cyclone forecasting.
\bibliography{main}

\newpage
\appendix

\setcounter{section}{0}
\setcounter{figure}{0} 
\setcounter{table}{0}
\setcounter{equation}{0}
\renewcommand{\thesection}{A.\arabic{section}}
\renewcommand{\thefigure}{A.\arabic{figure}}
\renewcommand{\thetable}{A.\arabic{table}}
\renewcommand{\theequation}{A.\arabic{equation}}


\section*{Appendix}

\section{Data} \label{sec:data}

The following descriptions of ERA5 and \hresfczero datasets are largely adapted from \citet{lam2022graphcast} and \citet{price2023gencast}. We include some additional details around preprocessing of HRES-fc0 for the purposes of finetuning \ourmodel.

\subsection{ERA5}

For developing \ourmodel, we built our "pre-training" dataset from a subset of ECMWF's ERA5~\citep{hersbach2020era5}\footnote{See ERA5 documentation: \url{https://confluence.ecmwf.int/display/CKB/ERA5}.} archive, which is a large corpus of data that represents the global weather from 1959 to the present, at 1 hour increments, for hundreds of static, surface, and atmospheric variables. The ERA5 archive is based on \textit{reanalysis}, which uses ECMWF's HRES model (cycle 42r1) that was operational for most of 2016, within ECMWF's 4D-Var data assimilation system. ERA5 assimilates 12-hour windows of observations, from 21z-09z and 09z-21z, as well as previous forecasts, into a dense representation of the weather's state, for each historical date and time.
The ERA5 archive consists of a deterministic reanalysis (ERA5) computed at \erafivenativedegree native latitude/longitude resolution.

Our ERA5 dataset contains a subset of available variables in ECMWF's ERA5 archive (\cref{tab:app:variables}), on 13 pressure levels\footnote{We follow common practice of using pressure as our vertical coordinate, instead of altitude. A ``pressure level'' is a field of altitudes with equal pressure. E.g., ``pressure level 500 \unit{hPa}'' corresponds to the field of altitudes for which the pressure is 500 \unit{hPa}. The relationship between pressure and altitude is determined by the geopotential variable.} corresponding to the levels of the WeatherBench \citep{rasp2020weatherbench} benchmark: 50, 100, 150, 200, 250, 300, 400, 500, 600, 700, 850, 925, 1000 \unit{hPa}.
The range of years included was 1979-01-01 to 2018-01-15, which were downsampled to 6-hour time intervals (corresponding to 00z, 06z, 12z and 18z each day). The downsampling is performed by subsampling, except for the total precipitation, which is accumulated for the 6 hours leading up to the corresponding downsampled time.

\paragraph{Variables with NaNs.}\label{sec:app:varsnan}
ERA5 sea surface temperature (SST) data contains NaNs over land by default. As preprocessing of the SST training and evaluation data, values over land are replaced with the minimum sea surface temperature seen globally in a subset of ERA5.

\begin{center}
\begin{table}[htbp]
\centering
\begin{tabular}{||c | c | c | c | c||} 
 \hline
 Type & Variable name & Short & ECMWF & Role (accumulation \\ [0.5ex] 
  &  &  name & Parameter ID & period, if applicable) \\ [0.5ex] 
 \hline\hline
 Atmospheric & Geopotential & z & 129 & Input/Predicted \\
\hline
Atmospheric & Specific humidity & q & 133 & Input/Predicted \\
\hline
Atmospheric & Temperature & t & 130 & Input/Predicted \\
\hline
Atmospheric & U component of wind & u & 131 & Input/Predicted \\
\hline
Atmospheric & V component of wind & v & 132 & Input/Predicted \\
\hline
Atmospheric & Vertical velocity & w & 135 & Input/Predicted \\
\hline
Single & 2 metre temperature & 2t & 167 & Input/Predicted 
\\
 \hline
Single & 10 metre u wind component & 10u & 165 & Input/Predicted \\
\hline
Single & 10 metre v wind component & 10v & 166 & Input/Predicted \\
\hline
Single & Mean sea level pressure & msl & 151 & Input/Predicted \\
 \hline
 Single & Sea Surface Temperature & sst & 34 & Input/Predicted \\
 \hline
Single & Total precipitation & tp & 228 & Predicted (6h) \\
 \hline
 \hline
\hline
Static & Geopotential at surface & z & 129 & Input \\
\hline
Static & Land-sea mask & lsm & 172 & Input \\
\hline
Static & Latitude & n/a & n/a & Input \\
\hline
Static & Longitude & n/a & n/a & Input \\
\hline
Clock & Local time of day & n/a & n/a & Input \\
\hline
Clock & Elapsed year progress & n/a & n/a & Input \\
 \hline
\end{tabular}
\caption{\small\textbf{ECMWF variables used in our datasets.} The ``Type'' column indicates whether the variable represents a \textit{static} property, a time-varying \textit{single}-level property (e.g., surface variables are included), or a time-varying \textit{atmospheric} property. The ``Variable name'' and ``Short name'' columns are ECMWF's labels. The ``ECMWF Parameter ID'' column is a ECMWF's numeric label, and can be used to construct the URL for ECMWF's description of the variable, by appending it as suffix to the following prefix, replacing ``ID'' with the numeric code: \texttt{https://apps.ecmwf.int/codes/grib/param-db/?id=ID}. The ``Role'' column indicates whether the variable is something our model takes as input and predicts, or only uses as input context (the double horizontal line separates predicted from input-only variables, to make the partitioning more visible).}
\label{tab:app:variables}
\end{table}
\end{center}

\subsection{\hresfczero}

For finetuning and evaluation, we make use of \hresfczero, which contains the initial (zeroeth) step of the ECMWF HRES forecast at each initialization time (00z, 06z, 12z, 18z) of the day. The HRES-fc0 data is similar to ERA5, but for a given forecast time, it is assimilated using the latest ECMWF NWP model, and only with observations from ±3 hours.

\paragraph{Variables with corresponding NaNs in ERA5.} Unlike ERA5, \hresfczero does not make use of NaNs to represent land within the sea surface temperature (SST) data. Rather, we found a placeholder value was used. To ensure a consistent representation between the two datasets in training, for all locations with a NaN in the original ERA5 dataset, we imputed the corresponding location in \hresfczero with the minimum sea surface temperature as well.

\paragraph{Total precipitation.} As an accumulated variable, total precipitation (\textit{tp}) takes a value of 0 in \hresfczero, inhibiting its use in a setting where the model is initialized on the dataset. To account for this, \ourmodel is trained to only output \textit{tp}, not taking it as input. Further, while ERA5 accumulates \textit{tp} over the interval leading up to the initialization time, using HRES \textit{tp} would require accumulating the variable forwards in time (since the interval prior is another forecast initialization's trajectory). To mitigate this discrepancy, we continue to use ERA5's \textit{tp} for both finetuning and evaluation purposes.

\section{Model Training}\label{sec:schedule}
The \ourmodel results reported in this paper were generated by a model trained in the following multi-stage process:
\begin{enumerate}[label={}]
\item \textbf{Stage 1:} 400000 steps with a \onedegree input and ground truth ERA5 dataset at a 12-hour temporal resolution.
\item \textbf{Stage 2:} 100000 steps with a \onedegree input and ground truth ERA5 dataset at a 6-hour temporal resolution.
\item \textbf{Stage 3:} 32000 steps with a \quarterdegree input and ground truth ERA5 dataset at a 6-hour temporal resolution.
\item \textbf{Stage 4:} 18000 autoregressive (AR) steps with a \quarterdegree ground truth \hresfczero dataset at a 6-hour temporal resolution. Including 8000 steps at 1AR, 4 000 steps at 2AR and 1000 steps at 3-8AR each.
\end{enumerate}
In all cases, the model is trained with a batch size of 64. A full list of hyper-parameters used for the stages are detailed in Table \ref{sec:schedule:table}.

We subsample the 6-hourly data with a stride of two to generate the 12-hourly dataset. Accumulated variables, such as total precipitation, are accumulated over the longer interval accordingly. Similarly, we downscale ERA5 from \quarterdegree to \onedegree by subsampling the spatial dimensions with a stride of four.

When training on \onedegree data, the graph-transformer processor operates on a 5-times refined icosahedral mesh. This is further refined a sixth time in the stages training on \quarterdegree data, where the architecture is updated to input and output the higher resolution data, but using the same model weights. To preserve the scale of the incoming signal at the start of Stage 3, we account for the increase in the number of number of messages being received by each mesh node and divide the sum of the message vectors by 4 in the Encoder GNN.

\begin{table}[htp] 
\centering
\begin{tabular}{|c|c|c|c|c|}
    \hline
    & \textbf{Stage 1} & \textbf{Stage 2} & \textbf{Stage 3} & \textbf{Stage 4} \\
    \hline
    \textbf{Training dataset} & ERA5 & ERA5 & ERA5 & \hresfczero \\
    \hline
    \textbf{Time-step size} & 12 hr & 6 hr & 6 hr & 6 hr \\
    \hline
    \textbf{Spatial resolution} & \onedegree & \onedegree & \quarterdegree & \quarterdegree \\
    \hline
    \textbf{Number of AR steps} & 1 & 1 & 1 & 1-8 \\
    \hline
    \textbf{Batch size} & \multicolumn{4}{c|}{64} \\
    \hline
    \textbf{Num training samples} & \multicolumn{4}{c|}{2} \\
    \hline
    \textbf{Optimizer} & \multicolumn{4}{c|}{AdamW \citep{loshchilov2018decoupled}} \\
    \hline
    \textbf{Weight decay} & \multicolumn{4}{c|}{0.1} \\
    \hline
    \textbf{LR decay schedule} & \multicolumn{4}{c|}{Cosine} \\
    \hline
    \textbf{Peak LR} & 8e-4 & 8e-5 & 8e-5 & 8e-5, 8e-6, 8e-7 thereafter \\
    \hline
    \textbf{Linear warm-up steps} & 1000 & 1000 & 1000 & 800 1AR, 400 2AR, 100 thereafter \\
    \hline
    \textbf{Total train steps} & 400 000 & 100 000 & 32 000 & 8 000 1AR, 4 000 2AR, 1 000 thereafter \\
    \hline
\end{tabular}
\caption{Model training hyperparameters.}
\end{table}
\label{sec:schedule:table}

\section{Architecture} \label{sec:app:architecture_changes}

\ourmodel uses a very similar architecture to that of \gencast, with some hyperparameter changes and some minor architectural differences.

\textbf{Conditioning on input physical states}. \gencast is a diffusion model, requiring multiple forward passes of the denoiser to generate a single forecast frame in a process of iterative refinement. As such, \gencast conditions on the two prior weather states, $X^{t-1, t-2}$, as well as a noisy version of the targets $Z_\sigma$, all concatenated along the channel dimension. \ourmodel generates a forecast in a single forward pass, and thus, similar to GraphCast \citep{lam2022graphcast} it simply conditions on the two prior weather states, $X^{t-1, t-2}$, without taking any noisy targets as inputs.

\textbf{Shared encoder for conditional layer-norm layers}. \gencast uses conditional layer-norm layers to condition on the noise level. First, the noise level is encoded into a vector of sine–cosine Fourier features, then it is passed through a small 2-layer MLP to obtain 16-dimensional noise-level encodings before ultimately being used to condition the various layer-norm layers in the architecture. \ourmodel also uses conditional layer-norm layers but to condition on the noise vector of 32 elements. In this case, the noise vector is directly encoded with a single matrix multiplication to produce another vector of 32 elements, and is then passed as conditioning to the various layer-norm layers. Initially this encoder was added as an implementation detail, but we later found that it seemed to help with optimization during training.

\textbf{Capacity}. \gencast contained MLPs with hidden and output sizes of 512, 4 attention heads, and 16 transformer layers. \ourmodel has a higher capacity with hidden and output sizes of 768, 6 attention heads, and 24 transformer layers. The only exception to this is the capacity of the MLPs that embed the edge features for the grid-to-mesh encoder graph and the mesh-to-grid decoder graph. For \gencast, this had hidden and output sizes of 512, while in \ourmodel these were reduced to 32. As there are $O(10^5)$ edges, but only 4 edge features to embed, we found this had no detrimental effect on performance while yielding significant savings in compute and memory.

\textbf{Global features}. \gencast (and GraphCast) condition on the sine and cosine of the of year progress to tell the model which time of the year is being modeled. They broadcast this feature, which is constant in space, across the entire grid, also providing it as input to the encoder in addition to other grid features. In \ourmodel, we additionally broadcast this feature across the entire mesh, concatenating it with the rest of the mesh input features. 

\textbf{Encoder from the grid to mesh}. \gencast (and GraphCast) use a graph neural network (GNN) to translate features between the grid and mesh structures by building a bipartite graph with edges pointing from the grid nodes to the mesh nodes. In these, the "edge model" (also often referred as "message function") is a function that conditions on both the features of the sender grid nodes, and the receiver mesh nodes, via gather operations, for every edge of the graph. For \ourmodel, we found that we could remove this conditioning on the receiver mesh nodes in the message function at no performance cost and significant compute and memory savings.

\section{Ablations}\label{sec:ablations}

\begin{figure}
    \centering
    \includegraphics[width=\textwidth]{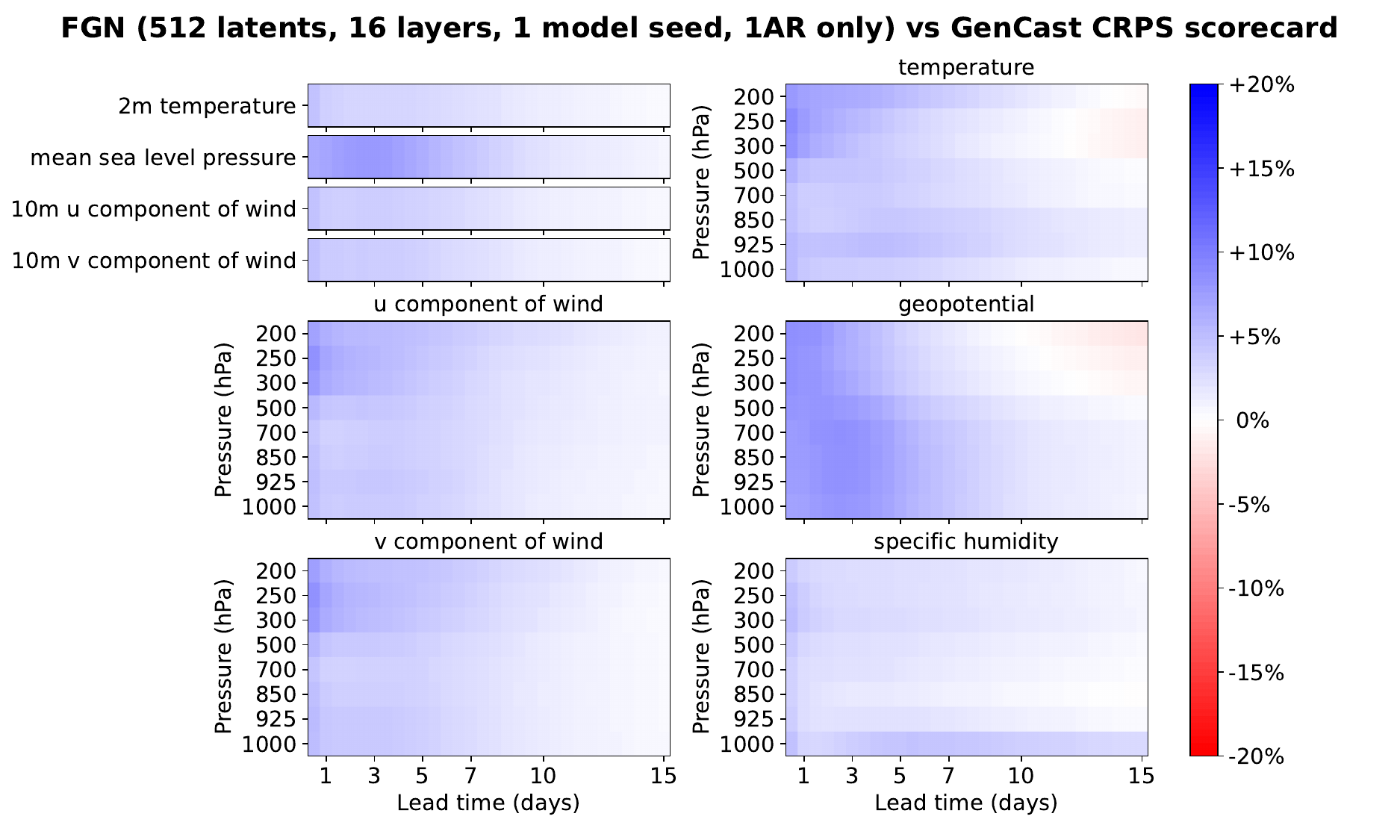}
    \caption{CRPS scorecard comparing \gencast with a version \ourmodel which has the same timestep (12 hours), same number of model seeds (1) and same model capacity (16-layers, and latent dimension of 512)---and which has also not undergone any autoregressive training. \ourmodel still matches or outperforms \gencast on globally averaged CRPS on 91\% of targets, with average improvement of 2.7\% and maximum improvement of 9\%, achieved on geopotential at shorter lead times.} \label{fig:sup:ablations_size_ensemble_ar}
\end{figure}

\begin{figure}
    \centering
    \includegraphics[width=\textwidth]{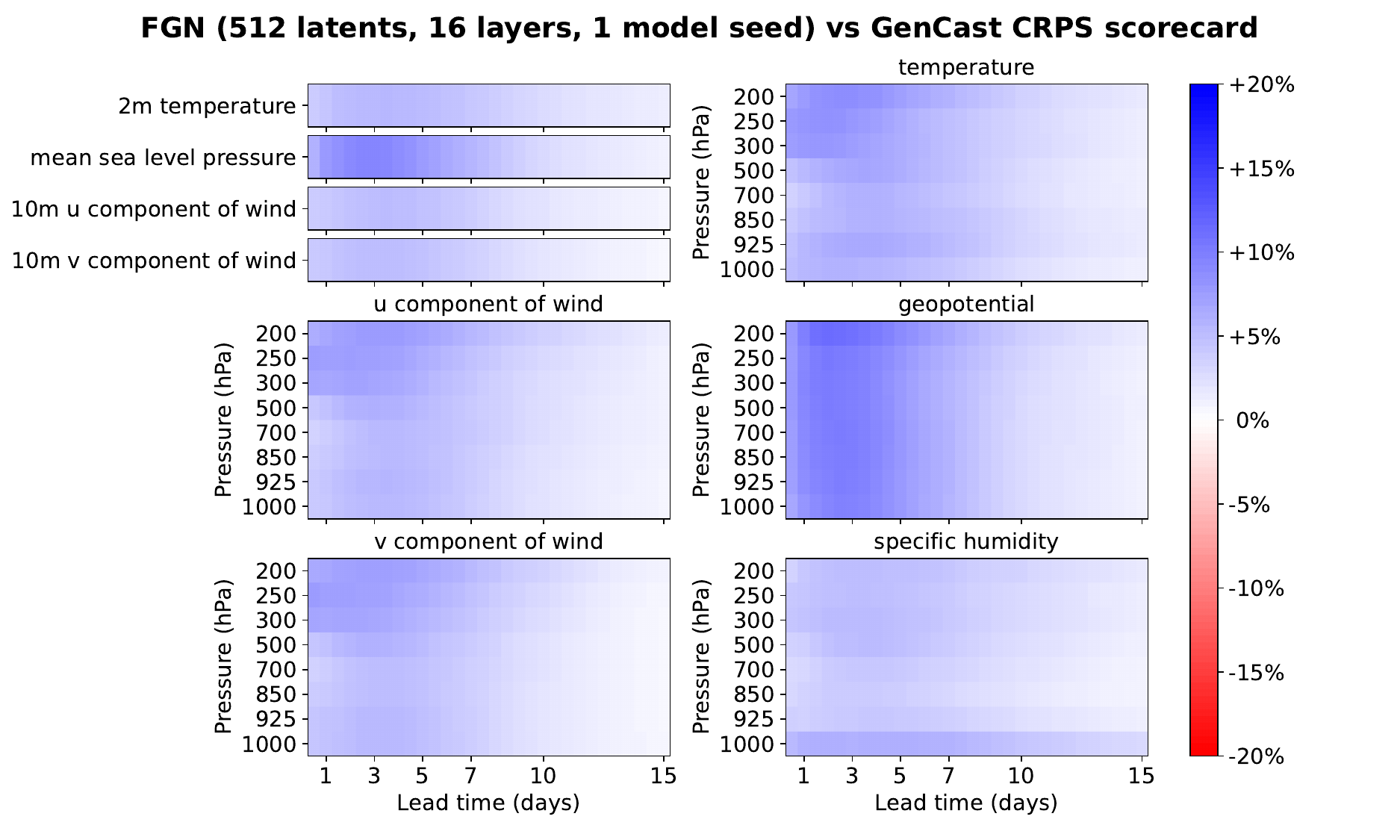}
    \caption{CRPS scorecard comparing \gencast with a version of \ourmodel which has the same timestep (12 hours), same number of model seeds (1) and same model capacity (16-layers, and latent dimension of 512)---but which has still undergone autoregressive training up to 8 rollout steps. This version of \ourmodel matches or outperforms \gencast on globally averaged CRPS on 99.8\% of targets, with average improvement of 4\% and maximum improvement of 11.5\%, achieved on geopotential at shorter lead times.}    \label{fig:sup:ablations_size_ensemble}
\end{figure}

Figure \ref{fig:sup:ablations_size_ensemble_ar} compares \gencast to a version of \ourmodel which has the same timestep (12 hours), same number of model seeds (1) and same model capacity (16-layers, and latent dimension of 512) as \gencast---and which has only been trained on a single step loss. This stripped-down version of \ourmodel still outperforms \gencast on more than 90\% of targets. Figure~\ref{fig:sup:ablations_size_ensemble} shows an \ourmodel model of the same capacity and timestep but which has also undergone autoregressive training up to 8 rollout steps. The results further improve, with this version of \ourmodel outperforming \gencast on CRPS on 99\% of targets and with average improvement over \gencast increasing from 2.7\% to 4\%.

\begin{figure}
    \centering
    \includegraphics[width=\textwidth]{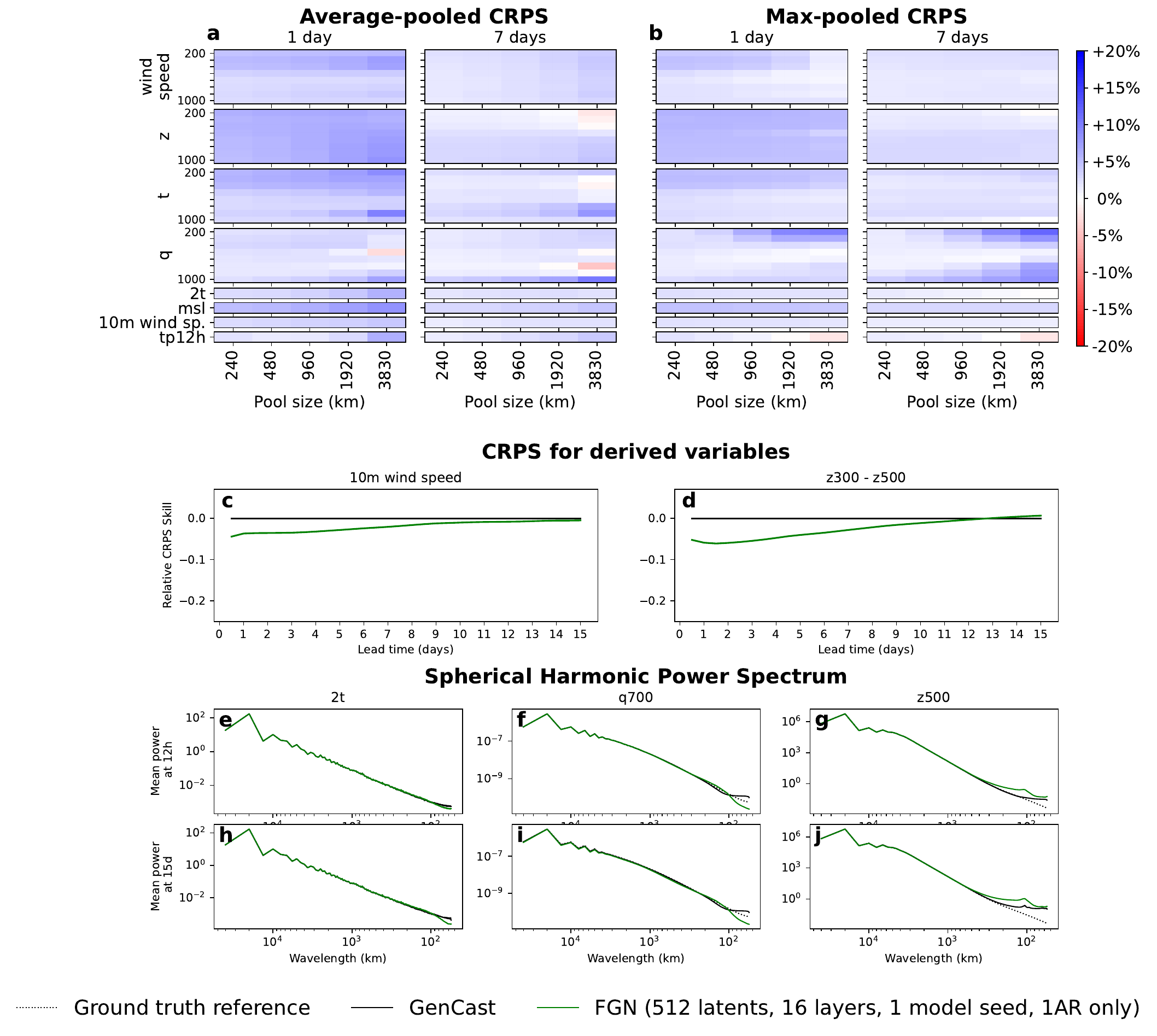}
    \caption{Evaluations of spatial and inter-variable structure in the joint forecast distributions, comparing \gencast with a version \ourmodel which has the same timestep (12 hours), same number of model seeds (1) and same model capacity (16-layers, and latent dimension of 512)---and which has also not undergone any autoregressive training.  
    (a, b) CRPS scorecards evaluating average-pool and max-pool fields respectively, for different pool sizes, at 1 and 7 day lead times.
    (c, d) Evaluations of skill on two quantities of interest derived from predicted fields, 10m wind speed and $z300 - z500$.
    (e - j) Power spectra for a number of variables.
    }    \label{fig:sup:joints_ablation}
\end{figure}

Figure~\ref{fig:sup:joints_ablation} reproduces Figure~\ref{fig:joints} but comparing \gencast to a version of \ourmodel which has the same timestep (12 hours), same number of model seeds (1), same model capacity (16-layers, and latent dimension of 512) as \gencast, and without auto-regressive training. The results show that even when controlling for model capacity, model ensemble size, and number of rollout steps during training, \ourmodel generally outperforms \gencast, successfully capturing spatial and inter-variable correlations. 

\begin{figure}
    \centering
    \includegraphics[width=\textwidth]{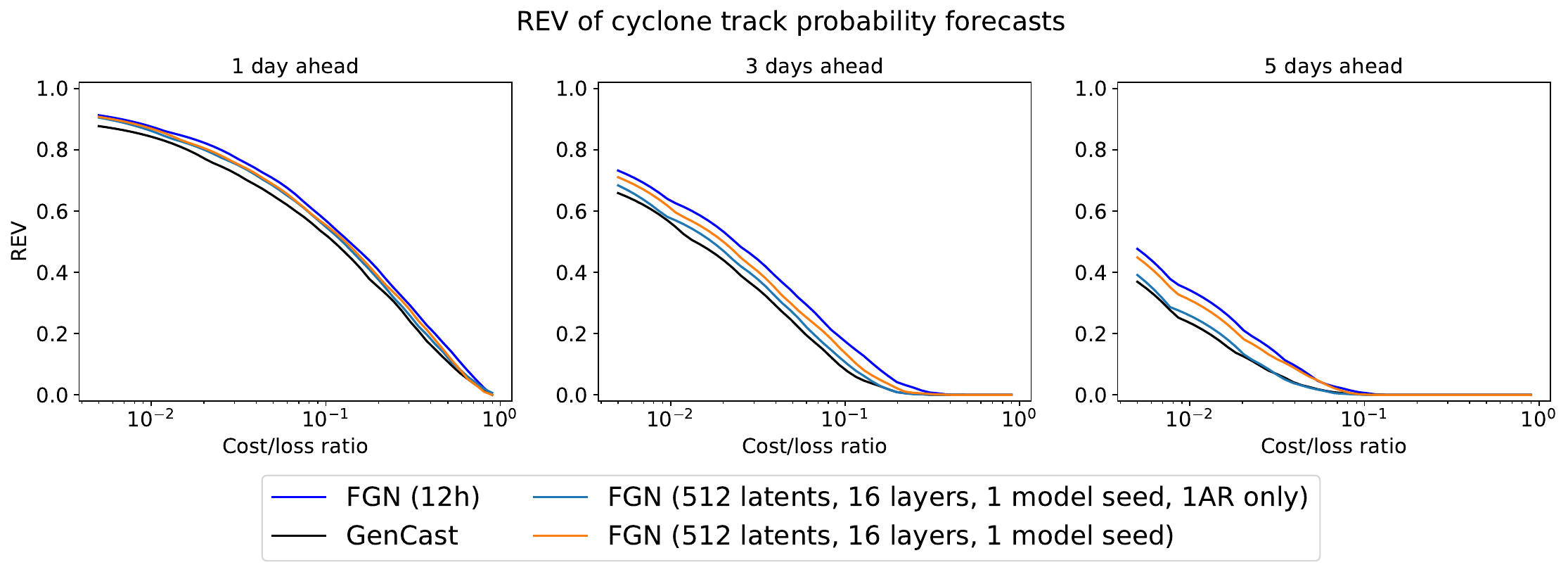}
    \caption{REV of cyclone track probability forecasts, comparing the impact of autoregressive training, model size, and model ensembling. The single-step trained \ourmodel version of the same depth and width as \gencast (green curve) performs as well as or better than \gencast. Adding autoregressive training (orange curve) further improves REV, and increasing the model size and using a model ensemble (blue) performs best.}
    \label{fig:sup:ablations_cyclones}
\end{figure}

Figure~\ref{fig:sup:ablations_cyclones} compares REV of the track probability predictions of \gencast to three models: the two smaller, 12-hour step, single seed versions of \ourmodel described in the previous paragraph, as well as a 12-hour step version of the full size, 4-model-seed \ourmodel model. The single-step trained \ourmodel version of the same size as \gencast (green curve) performs as well as or better than \gencast. Adding autoregressive training (orange curve) further improves REV, and increasing the model size and using a model ensemble (blue) performs best. 

Overall these results show that the simple CRPS-based training setup described in Section~\ref{sec:crps}---without model ensembling, model scaling, or autoregressive training---is sufficient to match or exceed SOTA forecast skill, but that the addition of these elements further boosts performance.

One drawback of the single-step trained \ourmodel is that, unlike in the 8AR step trained models, we observed some (rare) instances of unstable rollouts which correspond to forecasts with highly non-physical, artifact dominated forecasts states at late lead times. As an ad hoc investigation into this phenomena we visually inspected outlier forecasts as measured by any of the per-forecast mean absolute error, bias or L1 distance between the predictions and targets power spectra, for each variable for each level. This investigation suggested that the prevalence of this behavior was on the order of 0.1\% at 15-day lead-times (and smaller at shorter lead-times). Crucially we did not find any instances of this behavior for the autoregressively trained models (the ones evaluated in the main paper), which may suggest that the autoregressive training enables the models to be more robust and stable over rollouts.

\section{Supplementary Results and Figures}\label{sec:app:supplementary_results}

\subsection{Comparisons with ENS}\label{app:ens}

\begin{figure}
    \centering
    \includegraphics[width=\textwidth]{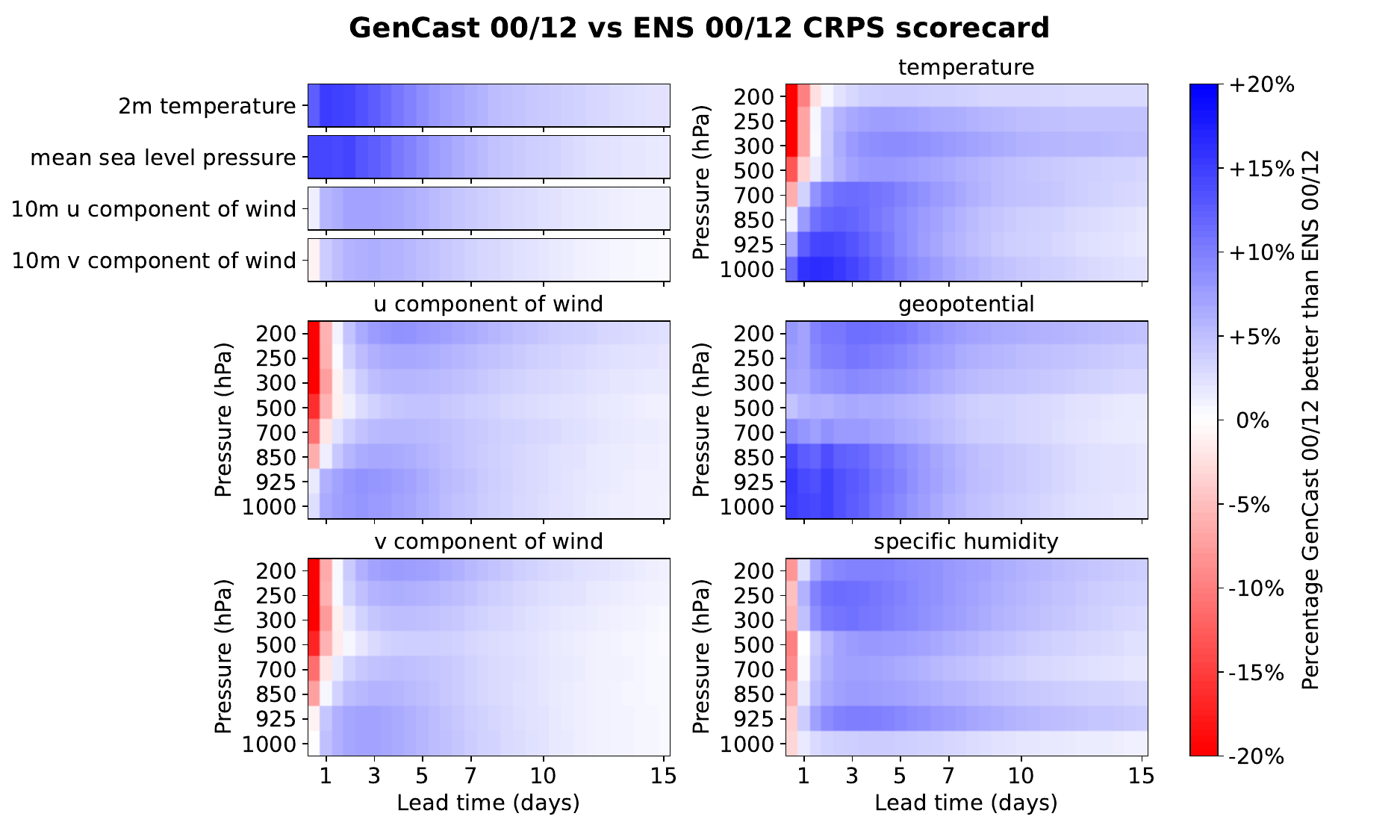}
    \caption{CRPS scorecard comparing the currently operational version of \gencast (used as the baseline in this work) to ENS on all 00/12 initialization times in 2023, against \hresfczero ground truth. \gencast outperforms ENS on 96.5\% of targets, with an average improvement of 4.7\% and a maximum improvement of 16.6\%. This confirms that the results established in \citet{price2023gencast} on ERA5 data generally carry over to the operational setting.}
    \label{fig:sup:gencast_ens}
\end{figure}

\citet{price2025probabilistic} established \gencast as the new state-of-the-art in probabilistic global weather forecasts, outperforming the prior state-of-the-art, ECMWF's ENS. As detailed in Section \ref{subsubsection:gencast}, this work makes use of a version of \gencast finetuned on HRES-fc0, whose operationally-initialized forecasts we compare to those of ENS over the year 2023. Figure \ref{fig:sup:gencast_ens} shows this \gencast model outperforms ENS on more than 96\% of targets, mirroring the results found in \citet{price2025probabilistic}, and confirming that \gencast is indeed the appropriate state-of-the-art baseline in the operational setting.

In these comparisons, we make use of the unbiased CRPS estimator to enable fair comparison, despite the fact that the \gencast ensemble is 56 members while ENS is 50 members. Moreover, as in \cite{price2025probabilistic}, we compare the models on 00z and 12z initializations as this is the data made available in the public TIGGE archive \citep{tiggearchive}.

\begin{figure}
    \centering
    \includegraphics[width=\textwidth]{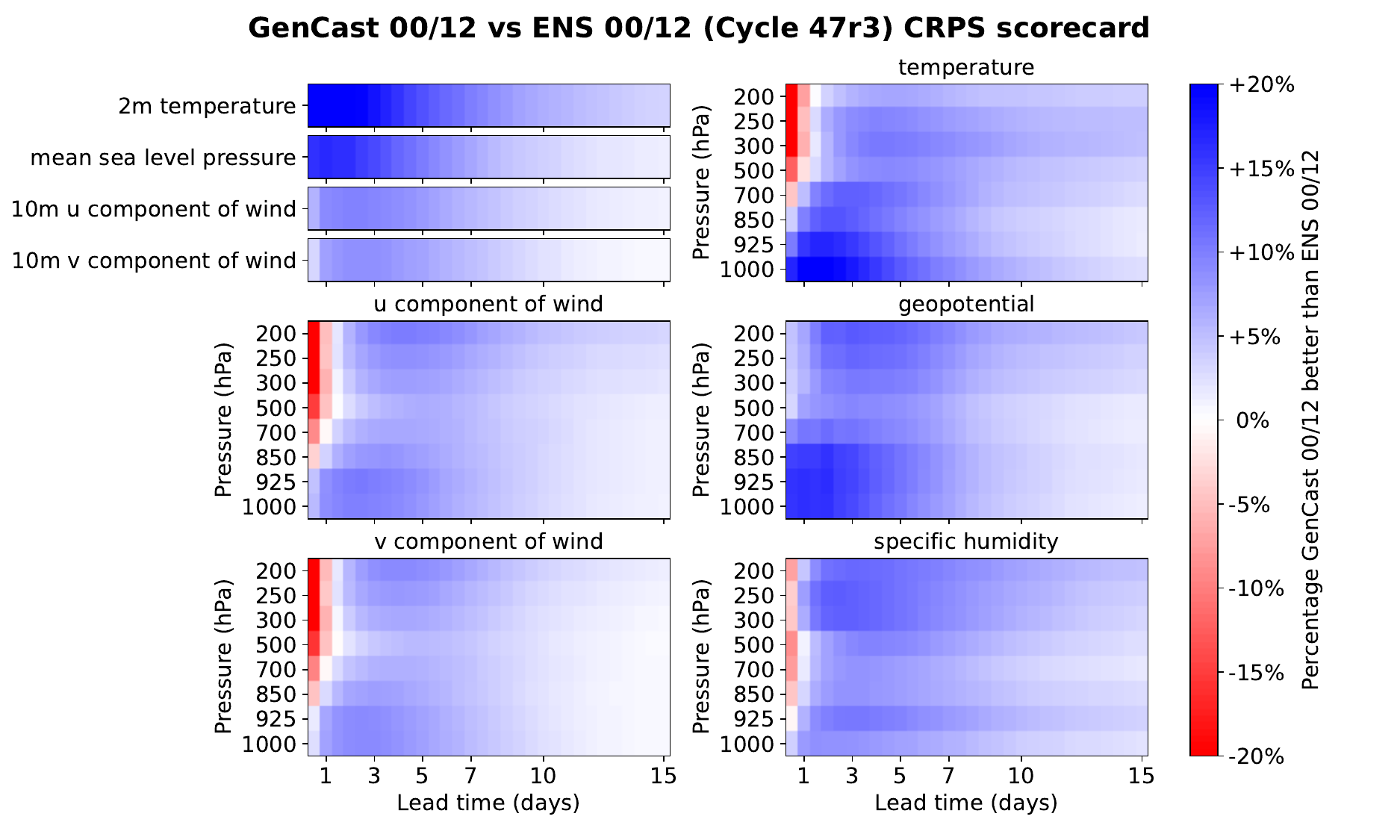}
    \caption{CRPS scorecard comparing the currently operational version of \gencast (used as the baseline in this work) to ENS on 00/12 initialization times before the 27th of June 2023, against \hresfczero ground truth. On these initializations, both models natively operate at resolutions (close to) \quarterdegree. \gencast outperforms ENS on 97.0\% of targets, with an average improvement of 5.6\% and a maximum improvement of 23.2\%.}
    \label{fig:sup:gencast_ens_cycle_47r3}
\end{figure}

\begin{figure}
    \centering
    \includegraphics[width=\textwidth]{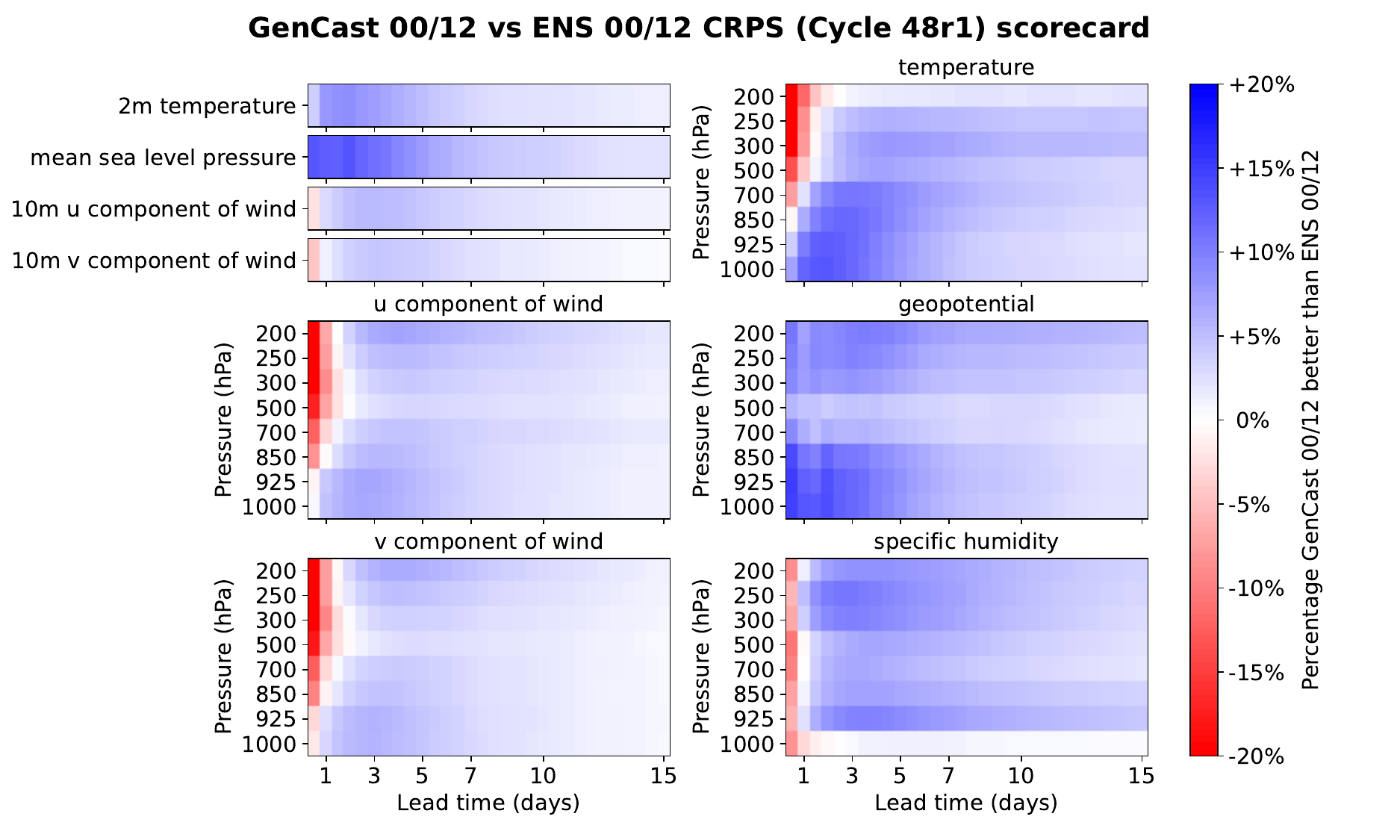}
    \caption{CRPS scorecard comparing the currently operational version of \gencast (used as the baseline in this work) to ENS on 00/12 initialization times after the 27th of June 2023, against \hresfczero ground truth. On these dates, ENS natively operates at a resolution \tenthdegree and is conservatively downsampled to \quarterdegree for this evaluation. \gencast outperforms ENS on 95.0\% of targets, with an average improvement of 3.7\% and a maximum improvement of 15.2\%.}
    \label{fig:sup:gencast_ens_cycle_48r1}
\end{figure}

We also note that on the 27th of June 2023, the resolution of ENS changed from \fifthdegree to \tenthdegree, as part of the upgrade to Cycle 48r1 from Cycle 47r3 \citep{cycle48r1}. To account for this, all data was regridded to \quarterdegree using MetView \citep{metview} defaults (conservative downsampling). Figures \ref{fig:sup:gencast_ens_cycle_47r3} and \ref{fig:sup:gencast_ens_cycle_48r1} present evaluations on dates in 2023 before and after this cycle change respectively, showing that while ENS improves in some variables after the upgrade, \gencast still outperforms ENS on 95\% of targets.

\begin{figure}
    \centering
    \includegraphics[width=\textwidth]{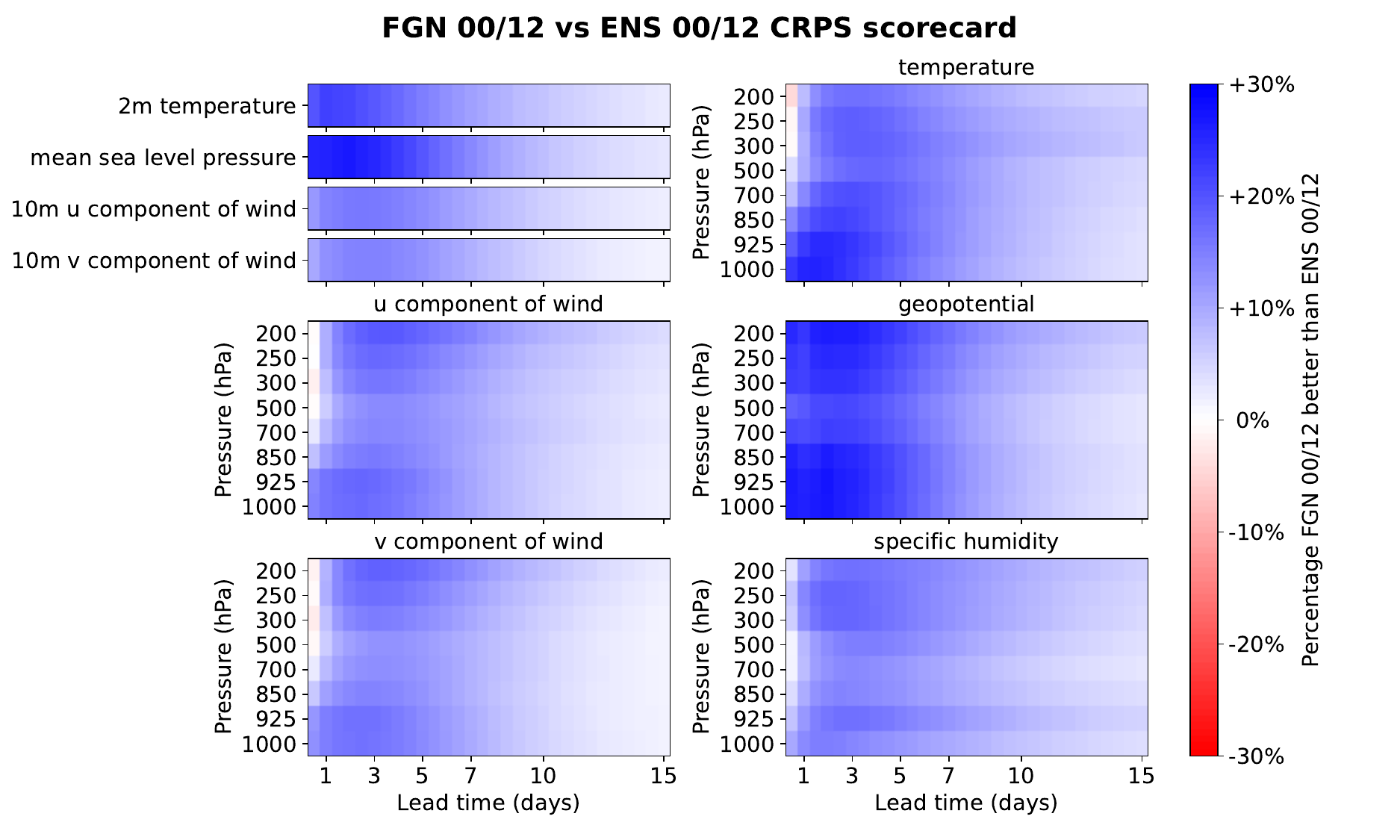}
    \caption{CRPS scorecard comparing \ourmodel to ENS, on all 00/12 initialization times in 2023, against \hresfczero ground truth (note the vlim of 30\%). \ourmodel achieves better CRPS on 99.3\% of targets, with an average improvement of 10.8\% and a maximum improvement of 27.7\%. Notably,  while \gencast suffered compared to ENS at short lead times, \ourmodel is strong across all lead times.}
    \label{fig:sup:fnv3_ens}
\end{figure}

\begin{figure}
    \centering
    \includegraphics[width=\textwidth]{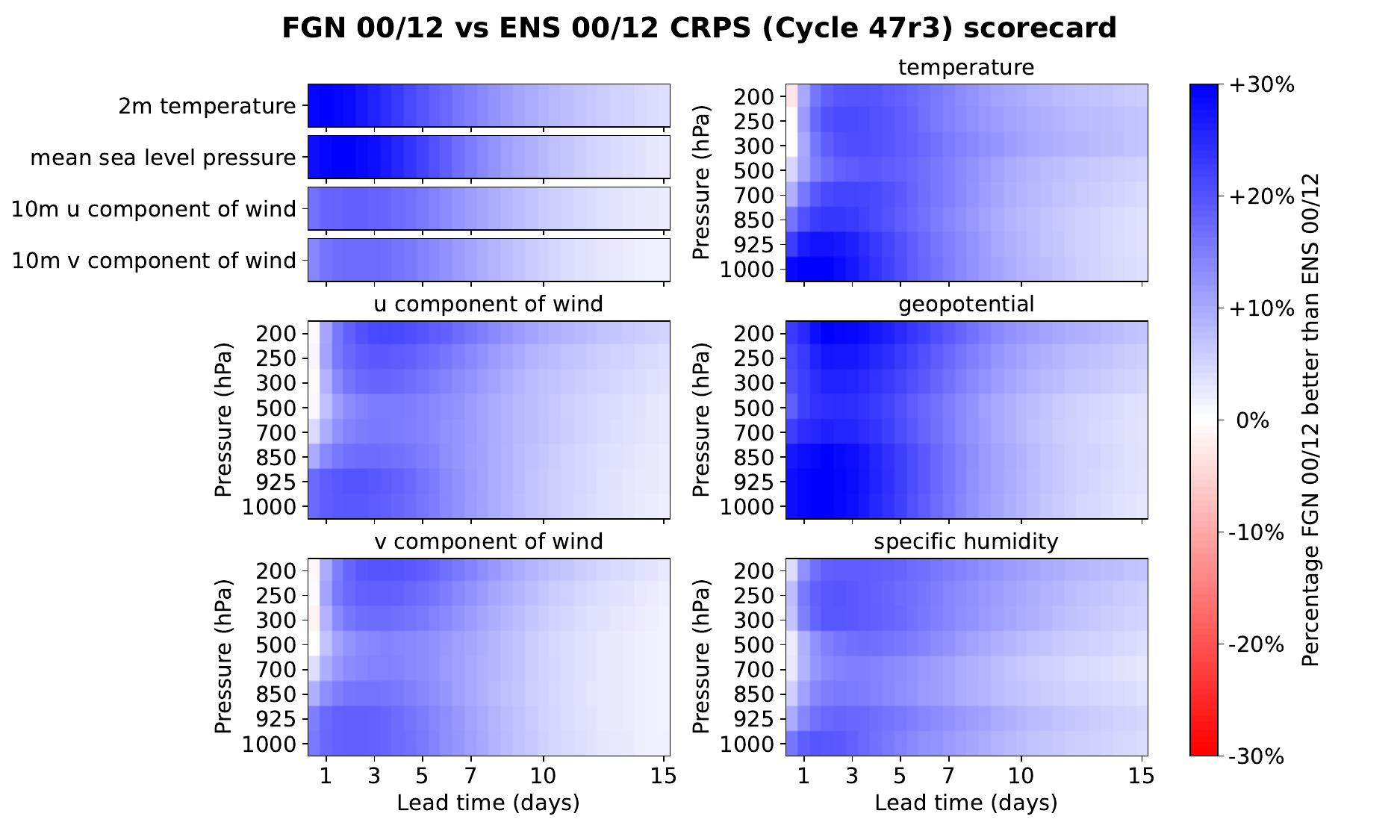}
    \caption{CRPS scorecard comparing \ourmodel to ENS, on 00/12 initialization times before the 27th of June 2023, against \hresfczero ground truth (note the vlim of 30\%). On these initializations, both models natively operate at resolutions (close to) \quarterdegree. \ourmodel achieves better CRPS on 99.5\% of targets, with an average improvement of 12.0\% and a maximum improvement of 30.6\%.}
    \label{fig:sup:fnv3_ens_cycle_47r3}
\end{figure}

\begin{figure}
    \centering
    \includegraphics[width=\textwidth]{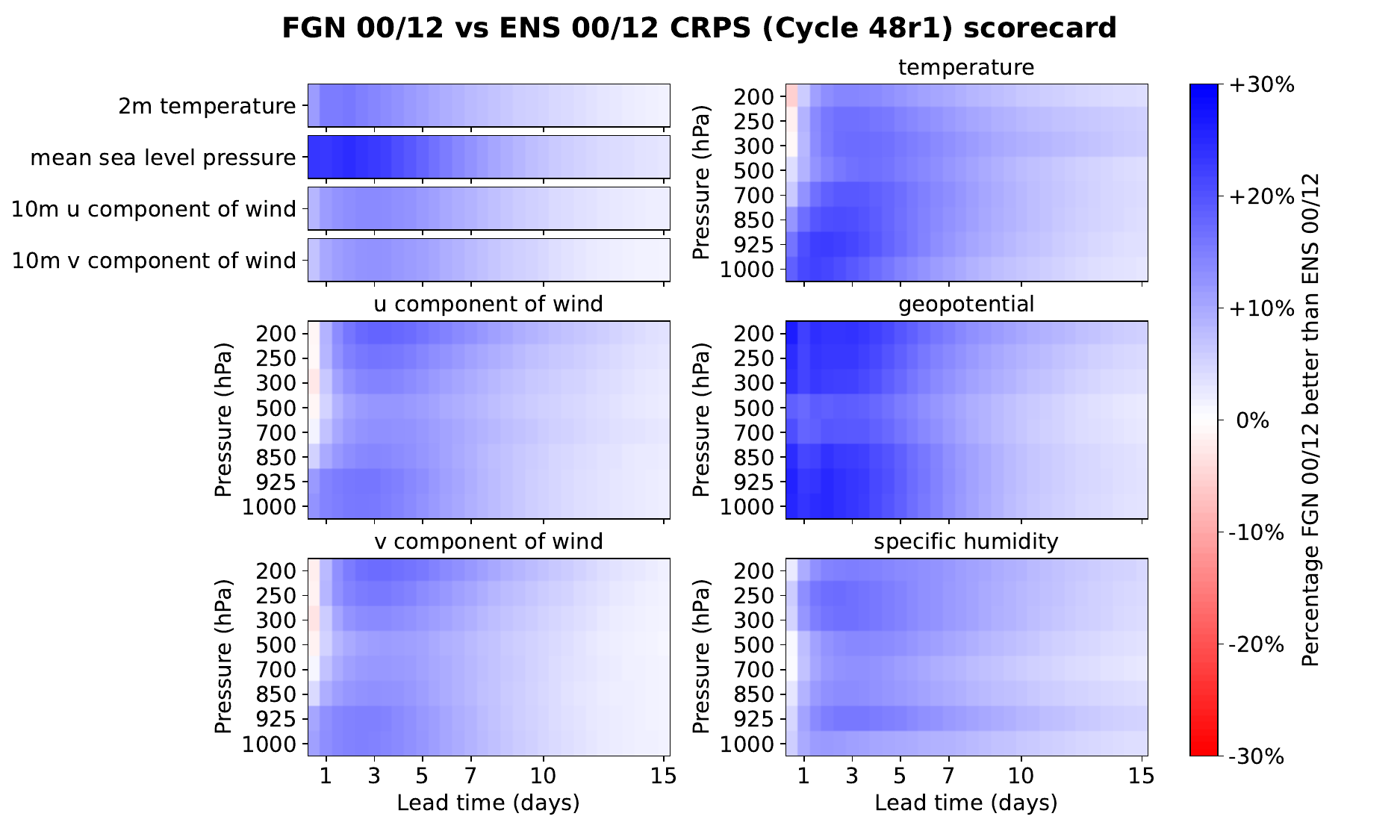}
    \caption{CRPS scorecard comparing ENS to \ourmodel on 00/12 initialization times, after the 27th of June 2023, against \hresfczero ground truth (note the vlim of 30\%). On these initializations, ENS natively operates at a resolution \tenthdegree and is conservatively downsampled to \quarterdegree for comparison. \ourmodel achieves better CRPS on 99.2\% of targets, with an average improvement of 9.7\% and a maximum improvement of 26.5\%.}
    \label{fig:sup:fnv3_ens_cycle_48r1}
\end{figure}

For completeness, Figure~\ref{fig:sup:fnv3_ens} compares \ourmodel to ENS, showing very large improvements (note the 30\% colorbar vlim), averaging more than 10\% across all lead times, ranging up to close to 30\%. Similarly, we report performance before and after the cycle change in Figures \ref{fig:sup:fnv3_ens_cycle_47r3} and \ref{fig:sup:fnv3_ens_cycle_48r1}, where \ourmodel outperforms ENS on more than 99\% of targets in both cases. In this comparison, technically neither the fair nor biased CRPS estimators or exactly appropriate - the \ourmodel ensemble being evaluated has 56 members while ENS has 50, but \ourmodel's ensemble also violates the independence assumption of the fair CRPS estimator when we enforce that each model seed generates an equal number of forecasts. \cref{fig:sup:fnv3_ens,fig:sup:fnv3_ens_cycle_47r3,fig:sup:fnv3_ens_cycle_48r1} report results using the fair estimator, which we found gives almost indistinguishable results from the biased setting, and so we are confident that these are representative.

\subsection{Extended Results}\label{sup:extended results}

\begin{figure}[p]
    \centering
    \includegraphics[width=\textwidth]{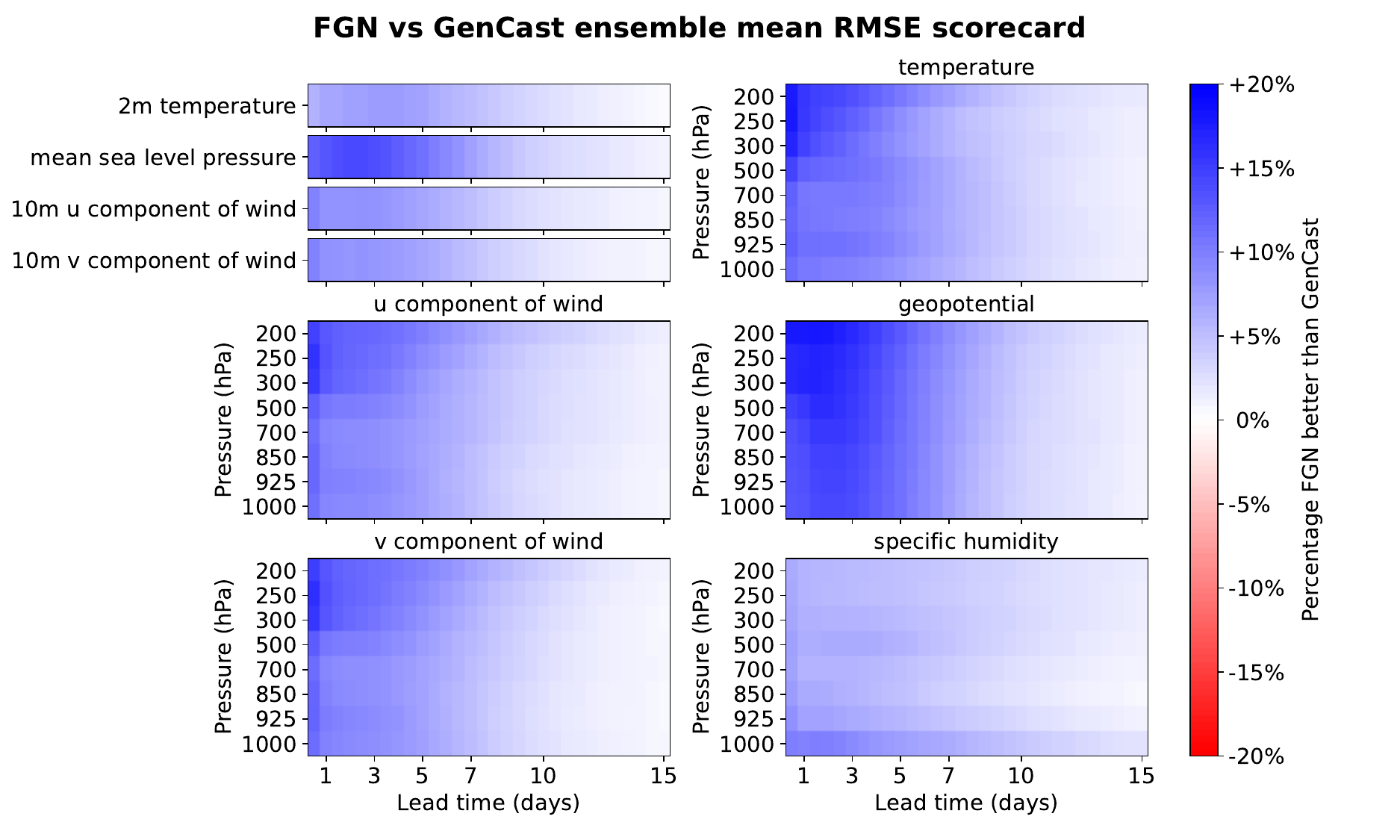}
    \caption{Ensemble mean RMSE scorecard comparing \ourmodel to \gencast, where blue cells indicate \ourmodel outperforms \gencast, red denotes where \gencast is better. \ourmodel is significantly better ($p < 0.05$) in 100\% of cases, with improvements of up to 18\% and an average of 5.8\%.}
    \label{fig:sup:rmse}
\end{figure}

\cref{fig:sup:rmse} shows the ensemble mean RMSE scorecards comparing \ourmodel to \gencast. 

\begin{figure}[p]
    \centering
    \includegraphics[width=\textwidth]{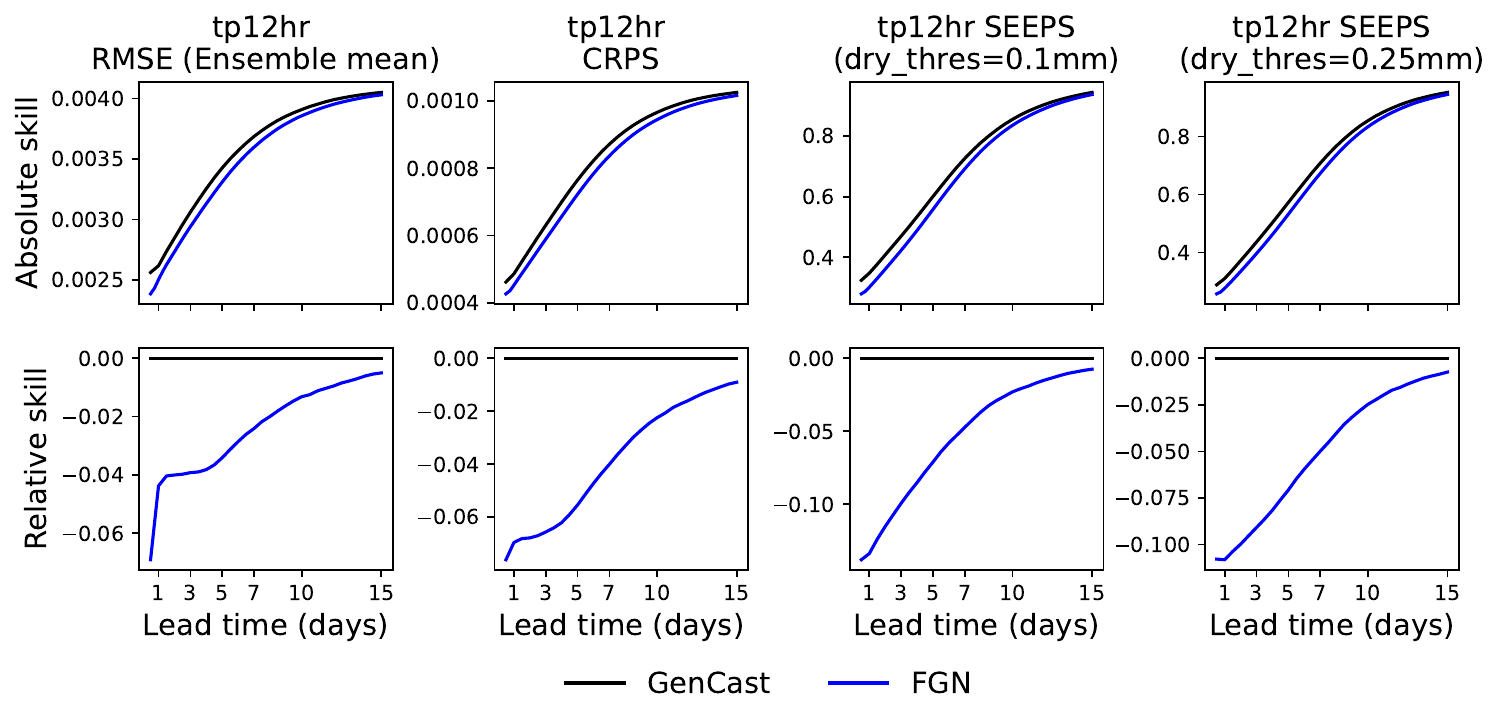}
    \caption{Results on 12h accumulated precipitation against ERA5 show \ourmodel outperforming \gencast when comparing Ensemble-Mean RMSE, CRPS and SEEPS (with dry thresholds of 0.1 and 0.25) for the predicted total accumulated precipitation over 12 hours. All metrics were calculated globally and over the full test period, except for SEEPS which excludes very dry regions according to the criteria in \citet{lam2023learning}. SEEPS is evaluated on the ensemble mean prediction. Absolute and relative plots are shown in the top and bottom rows respectively.}
    \label{fig:sup:precip_12}
\end{figure}

\begin{figure}[htbp]
    \centering
    \includegraphics[width=\textwidth]{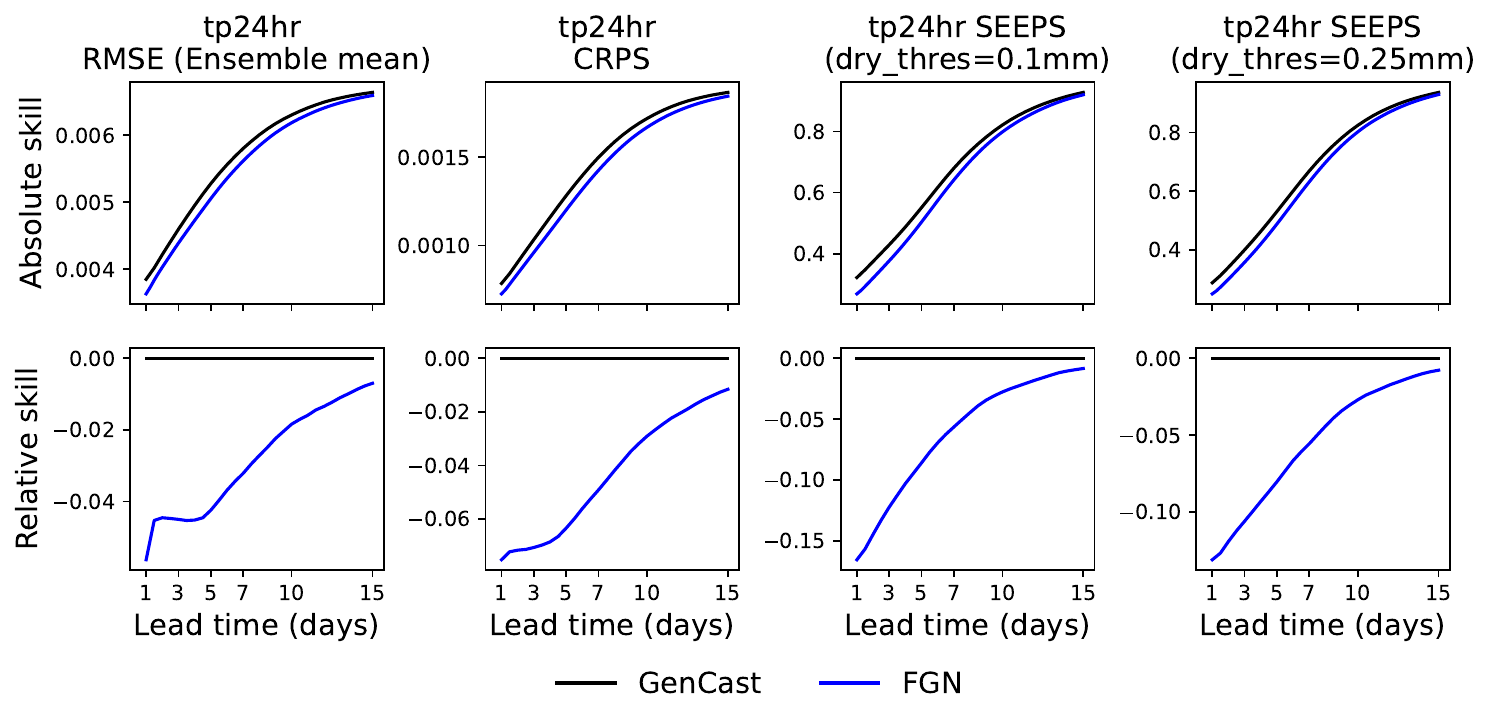}
    \caption{Similar to \cref{fig:sup:precip_12}, results on 24h accumulated precipitation against ERA5 also show \ourmodel outperforming \gencast.}
    \label{fig:sup:precip_24}
\end{figure}

We evaluate precipitation forecasting skill using RMSE and CRPS as well as  Stable Equitable Error in Probability Space (SEEPS), which is often used as a specialized metric for precipitation evaluation \citep{rodwell2010new,haiden2012intercomparison,north2013assessment}, using the same methodology as in \citet{lam2023learning} and \citet{price2025probabilistic}. Because SEEPS is a metric for deterministic categorical forecasts (dry, light-rain, and heavy-rain) we evaluated the ensembles computing the category of the ensemble mean.
Results for 12-hour and 24-hour accumulated precipitation are shown in \cref{fig:sup:precip_12}  and \cref{fig:sup:precip_24} respectively.

\begin{figure}
    \centering
    \includegraphics[width=\textwidth]{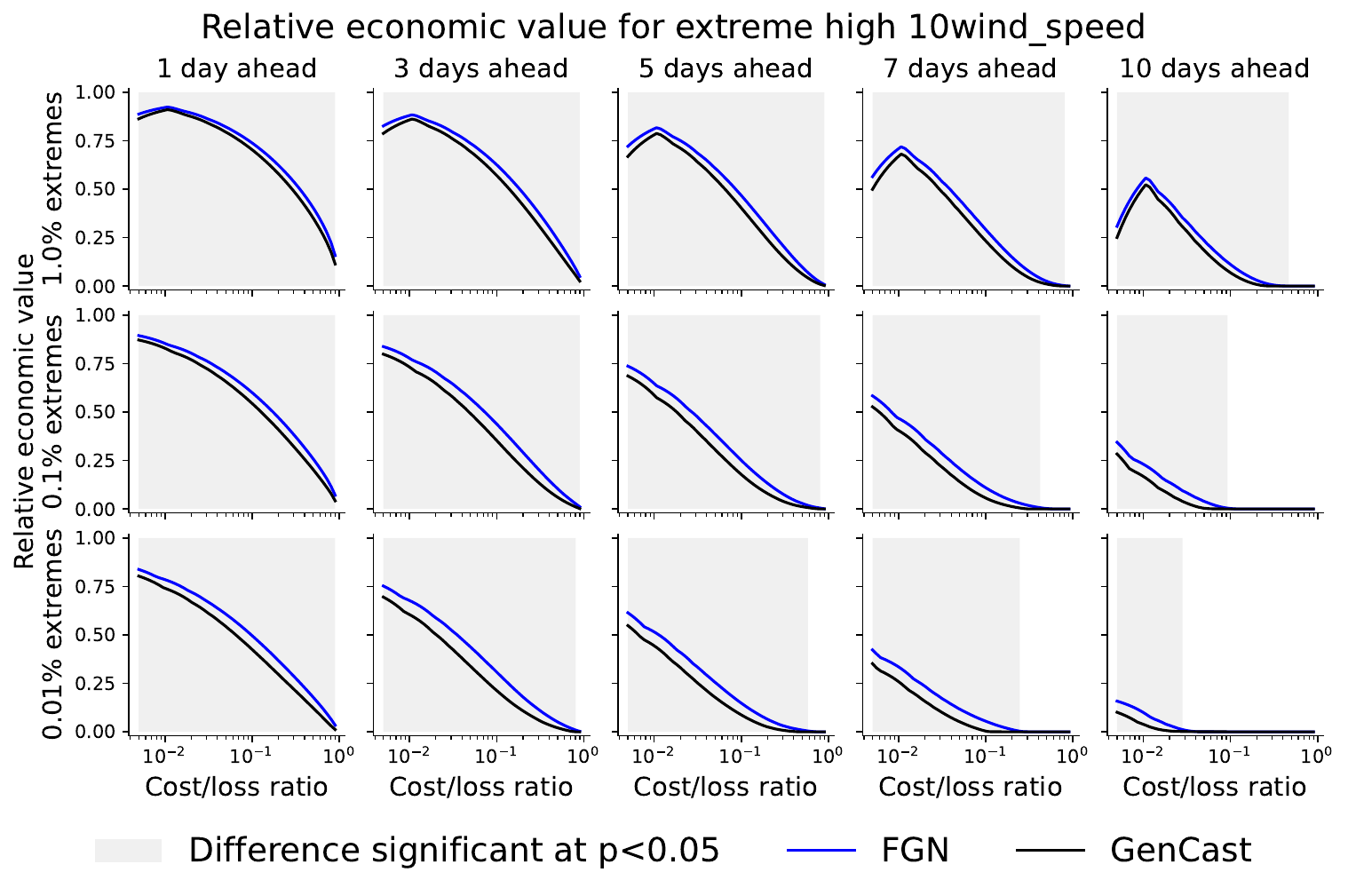}
    \caption{REV of predictions of extreme high 10m wind speed, across multiple cost/loss ratios, exceedance thresholds, and lead times. Cost/loss regions in which the difference between \ourmodel and \gencast is statistically significant ($p < 0.05$) are shaded in gray.}
    \label{fig:sup:rev_high_wind}
\end{figure}

\begin{figure}
    \centering
    \includegraphics[width=\textwidth]{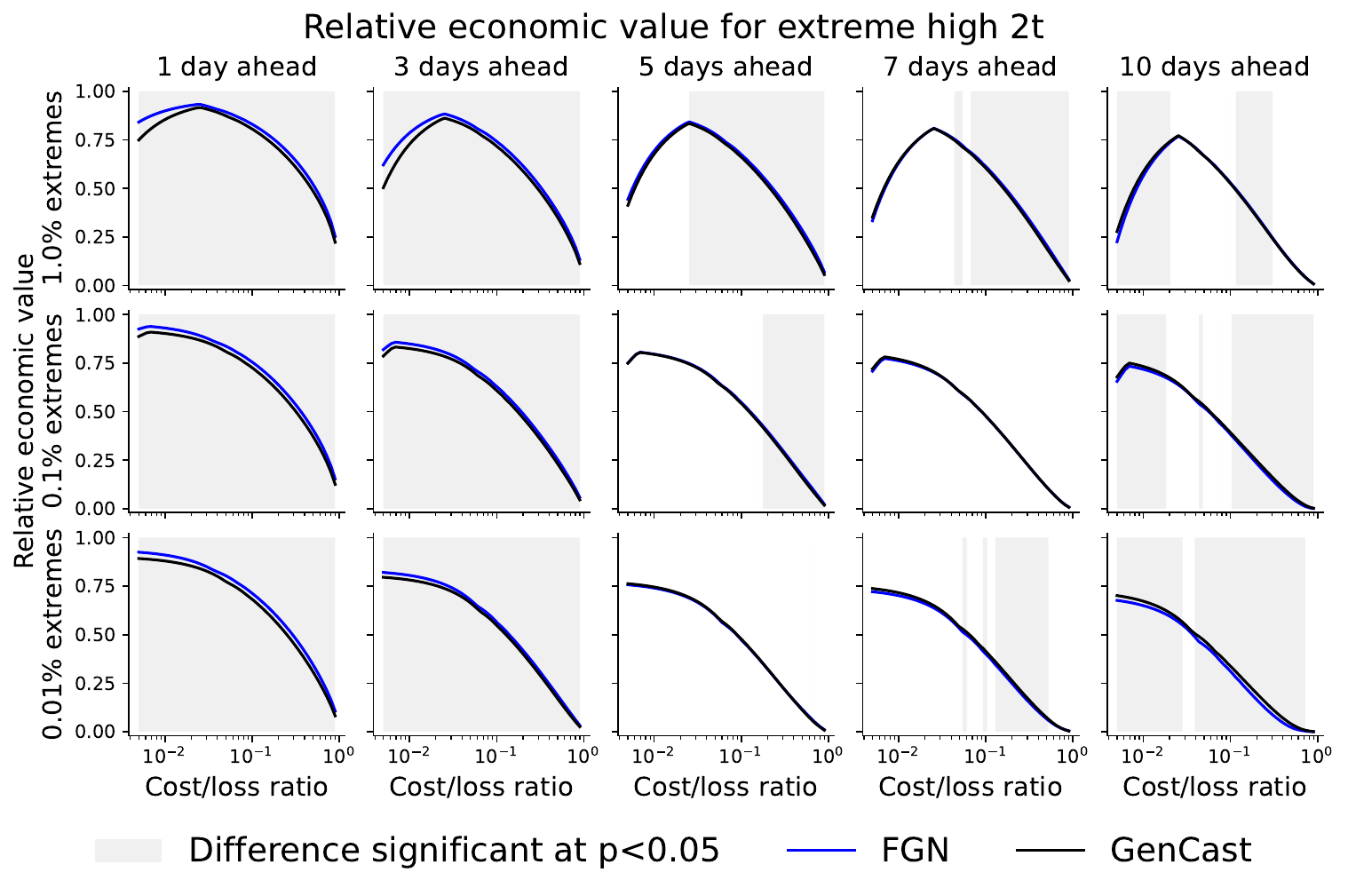}
    \caption{REV of predictions of extreme high 2m temperature, across multiple cost/loss ratios, exceedance thresholds, and lead times. Cost/loss regions in which the difference between \ourmodel and \gencast is statistically significant ($p < 0.05$) are shaded in gray.}
    \label{fig:sup:rev_high_2t}
\end{figure}

\begin{figure}
    \centering
    \includegraphics[width=\textwidth]{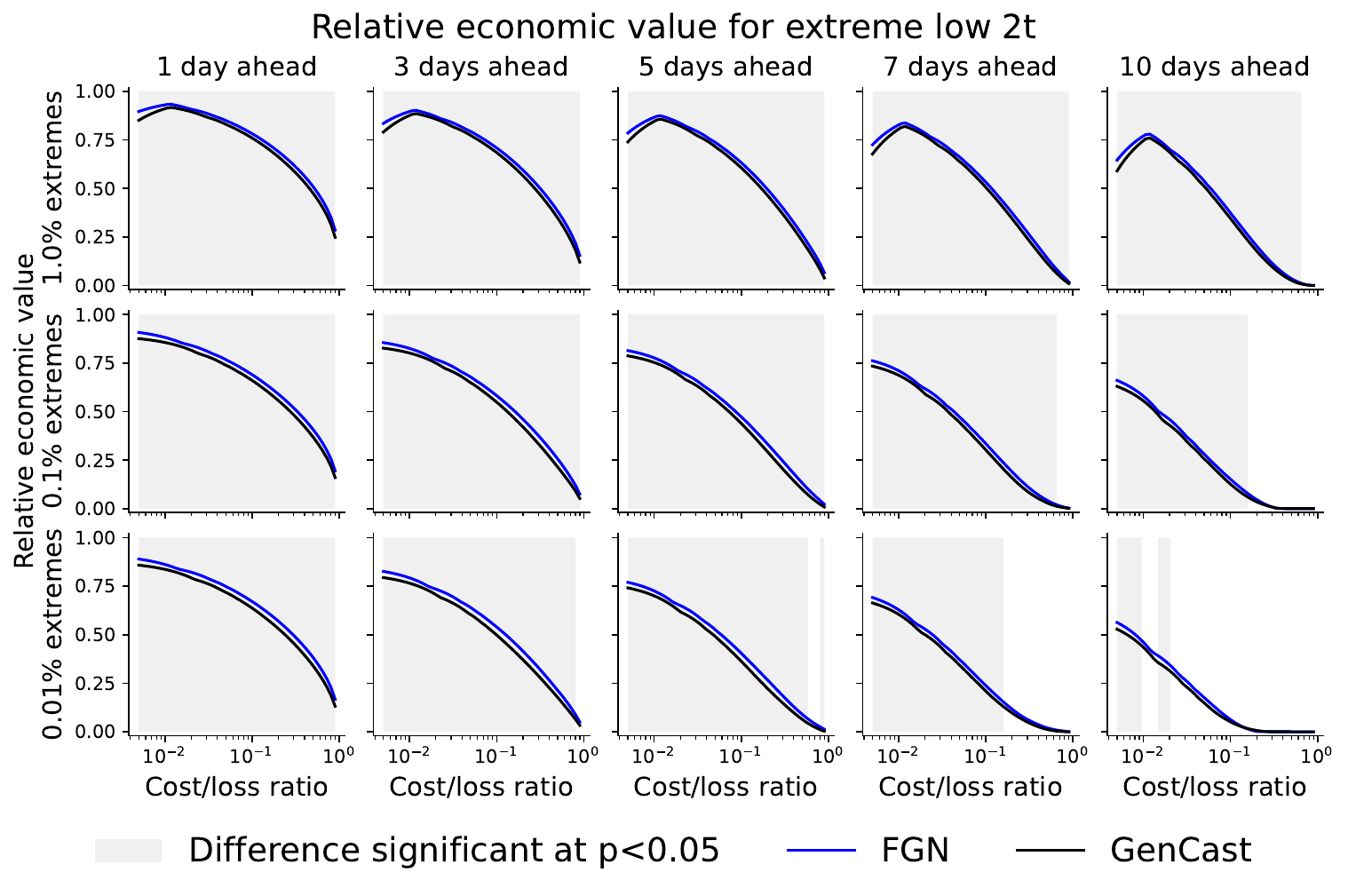}
    \caption{REV of predictions of extreme low 2m temperature, across multiple cost/loss ratios, exceedance thresholds, and lead times. Cost/loss regions in which the difference between \ourmodel and \gencast is statistically significant ($p < 0.05$) are shaded in gray.}
    \label{fig:sup:rev_low_2t}
\end{figure}

\begin{figure}
    \centering
    \includegraphics[width=\textwidth]{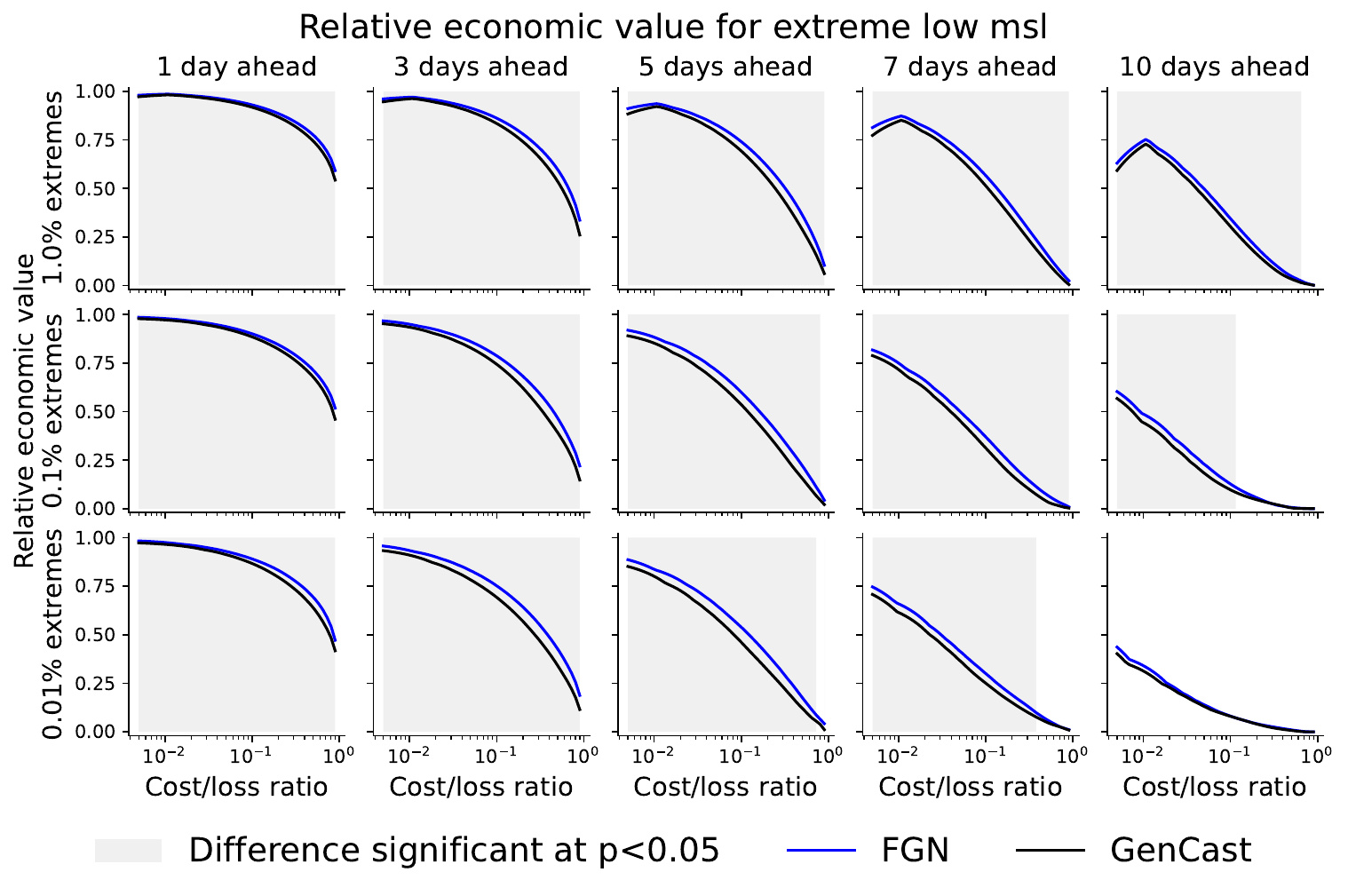}
    \caption{REV of predictions of extreme low mean sea level pressure, across multiple cost/loss ratios, exceedance thresholds, and lead times. Cost/loss regions in which the difference between \ourmodel and \gencast is statistically significant ($p < 0.05$) are shaded in gray.}
    \label{fig:sup:rev_low_msl}
\end{figure}

\cref{fig:sup:rev_high_wind,fig:sup:rev_high_2t,fig:sup:rev_low_2t}, and \ref{fig:sup:rev_low_msl} show REV results on prediction of extreme high 10m wind speed, high 2m temperature, low 2m temperature, and low mean sea level pressure, respectively, for different extreme thresholds, across different cost/loss ratios, for different lead times. \ourmodel achieves significantly higher REV than \gencast ($p < 0.05$), with a few exceptions at longer lead times for certain  cost-loss and lead time combinations for high 2m temperature.

\begin{figure}
    \centering
    \includegraphics[width=\textwidth]{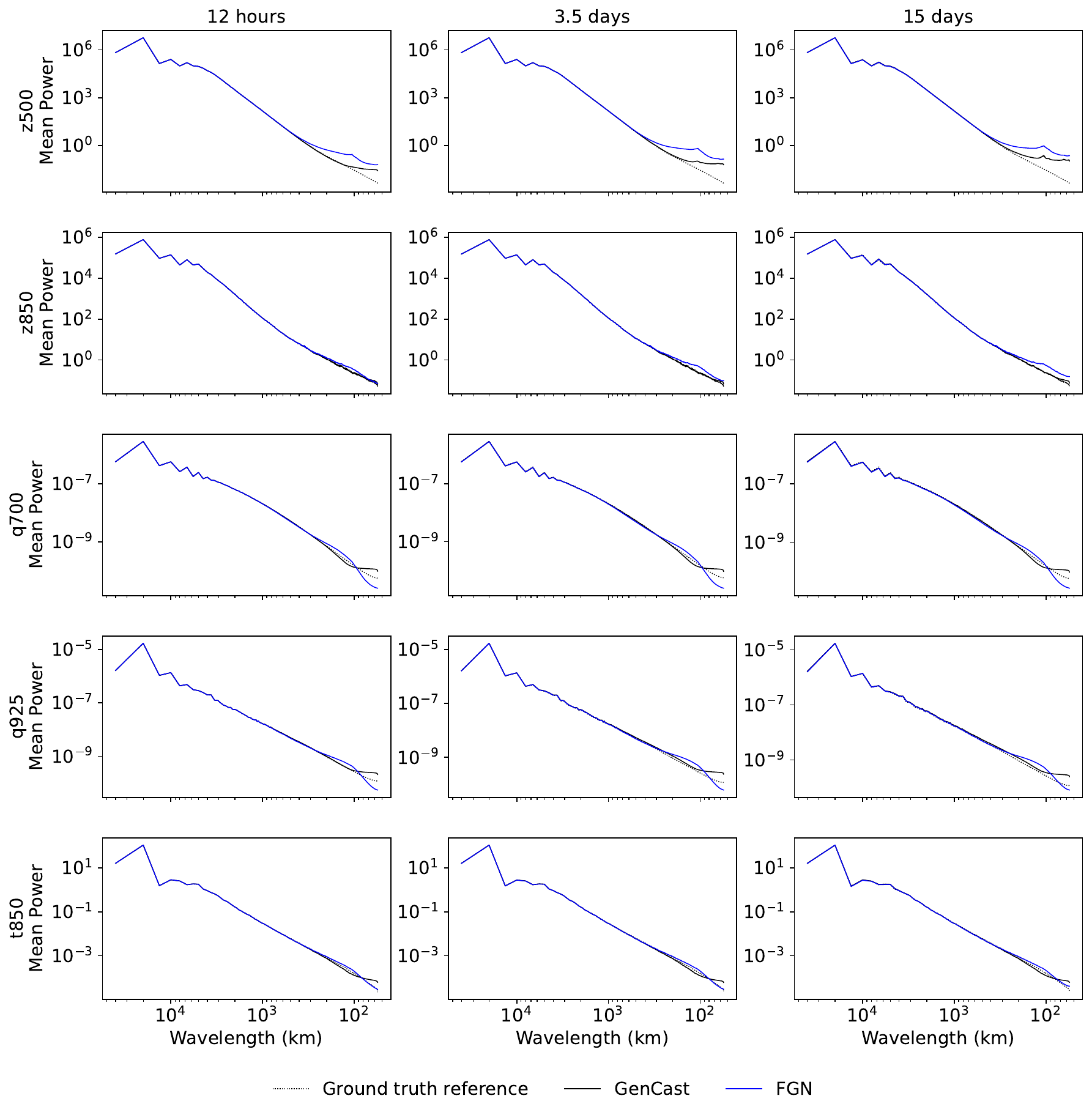}
    \caption{Power spectrum plots for z500, z850, q700, q925 and t850.}
    \label{fig:sup:spectra1}
\end{figure}

\begin{figure}
    \centering
    \includegraphics[width=\textwidth]{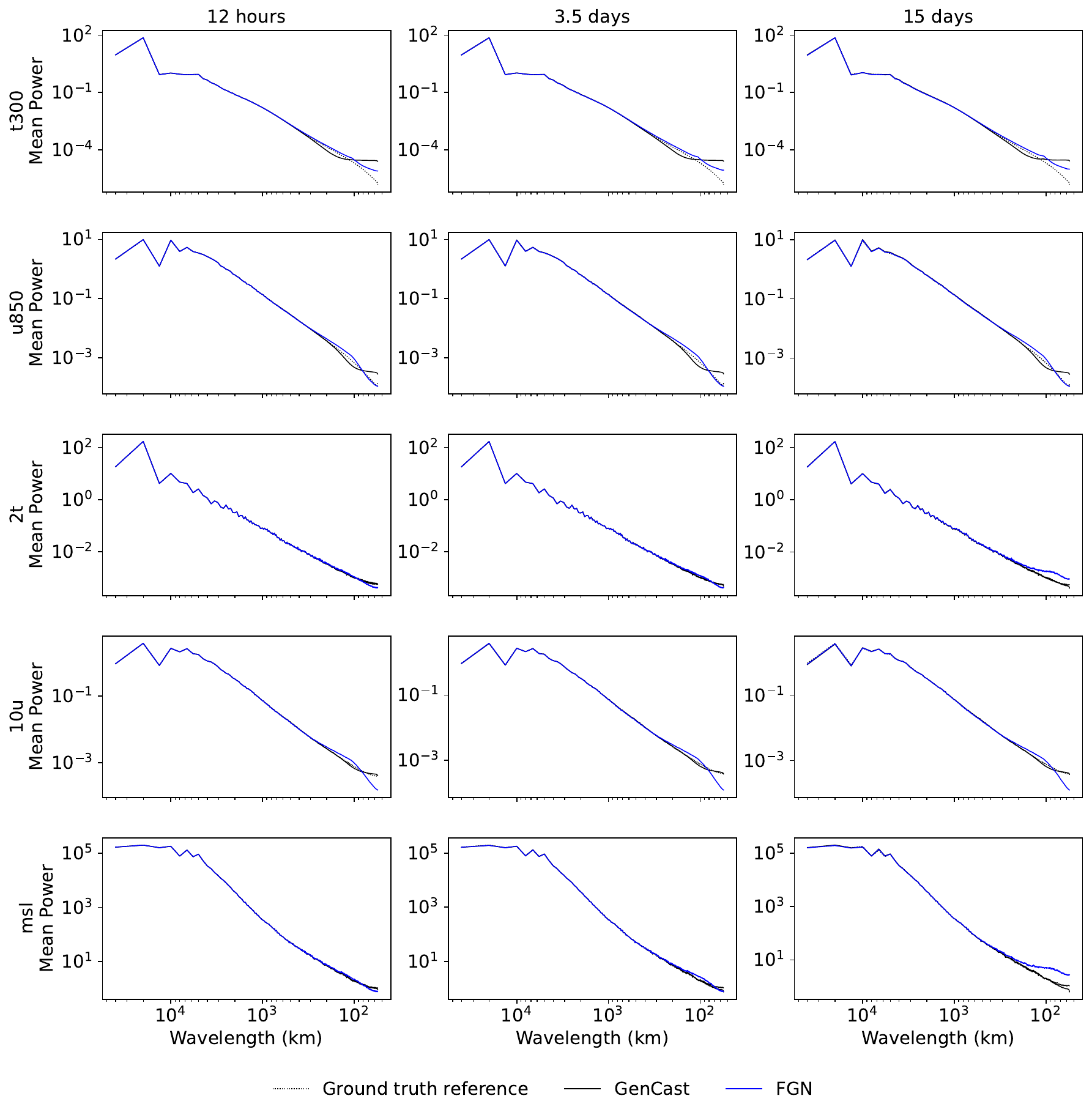}
    \caption{Power spectrum plots for t300, u850, 2t, 10u and msl.}
    \label{fig:sup:spectra2}
\end{figure}

We provide additional spectral results to supplement those the main paper, showing 10 representative variables: z500, z850, q700, q925, t850, t300, u850, 2t, 10u and msl (\cref{fig:sup:spectra1} and \cref{fig:sup:spectra2}).

\FloatBarrier

\FloatBarrier

\section{Forecast visualizations}\label{sec:app:visualizations}

\begin{figure}
    \centering
    \includegraphics[width=0.9\textwidth]{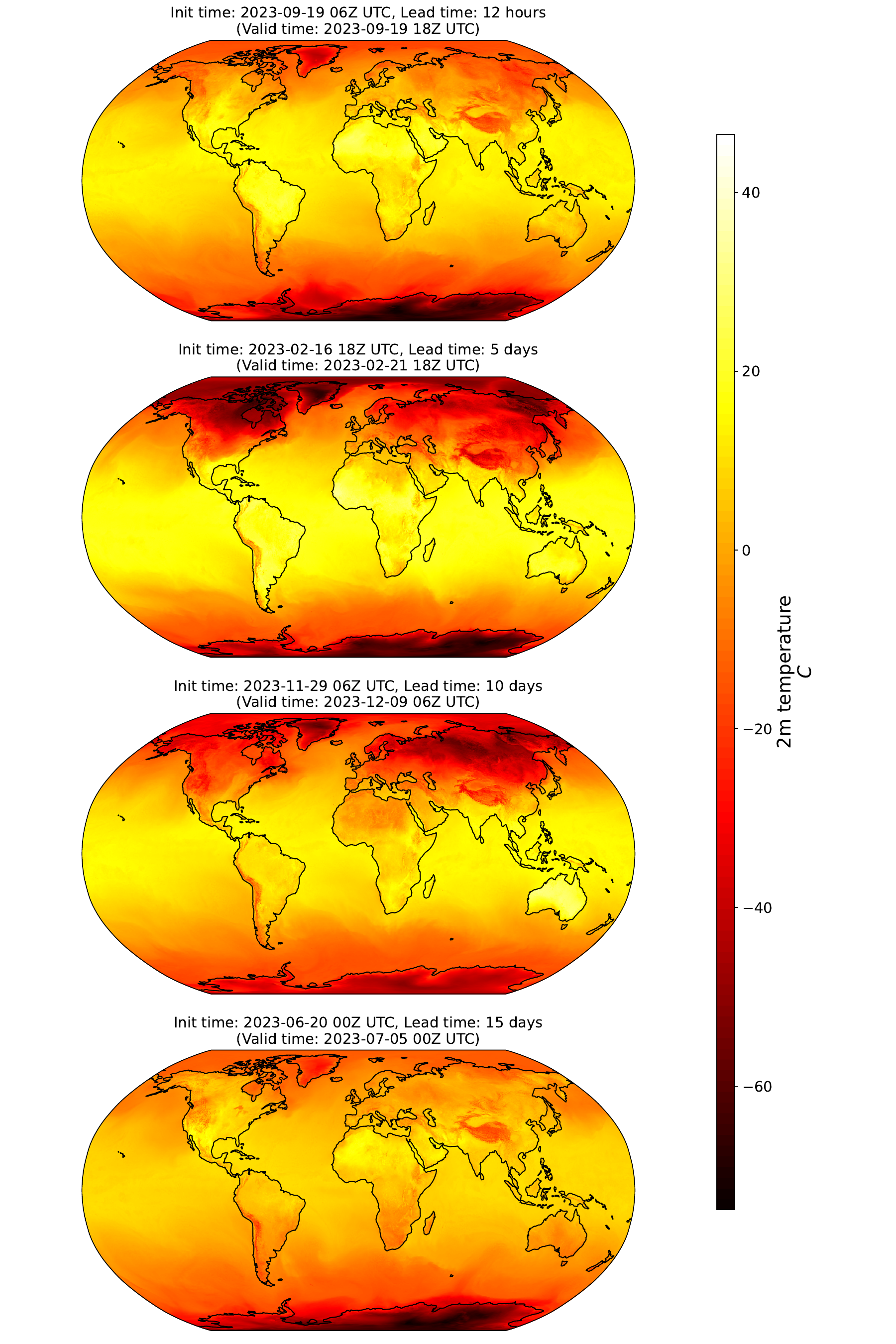}
    \caption{Visualization of 2m temperature.}
    \label{fig:sup:vis_2t}
\end{figure}

\begin{figure}
    \centering
    \includegraphics[width=0.9\textwidth]{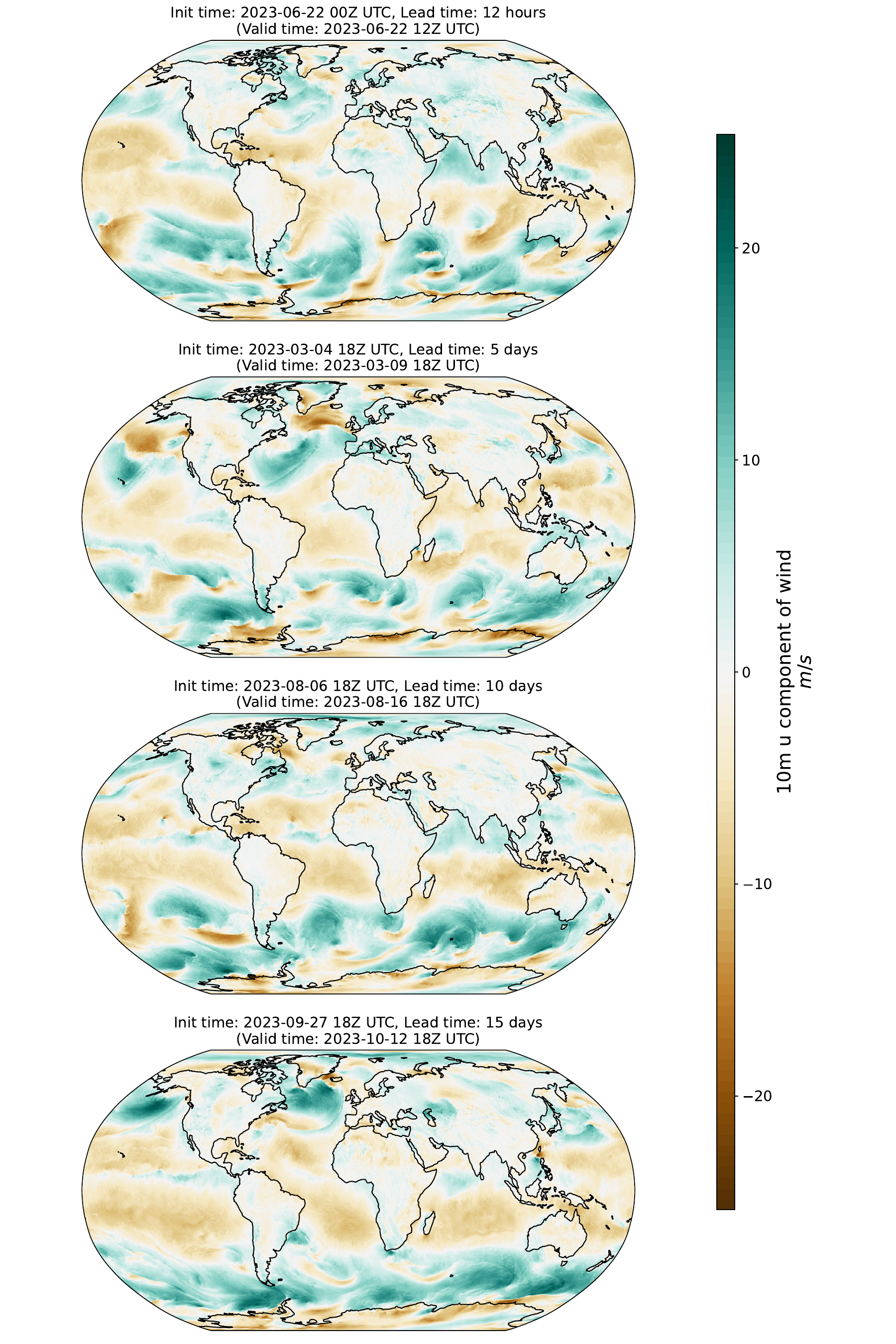}
    \caption{Visualization of 10 m u component of wind.}
    \label{fig:sup:vis_10u}
\end{figure}

\begin{figure}
    \centering
    \includegraphics[width=0.9\textwidth]{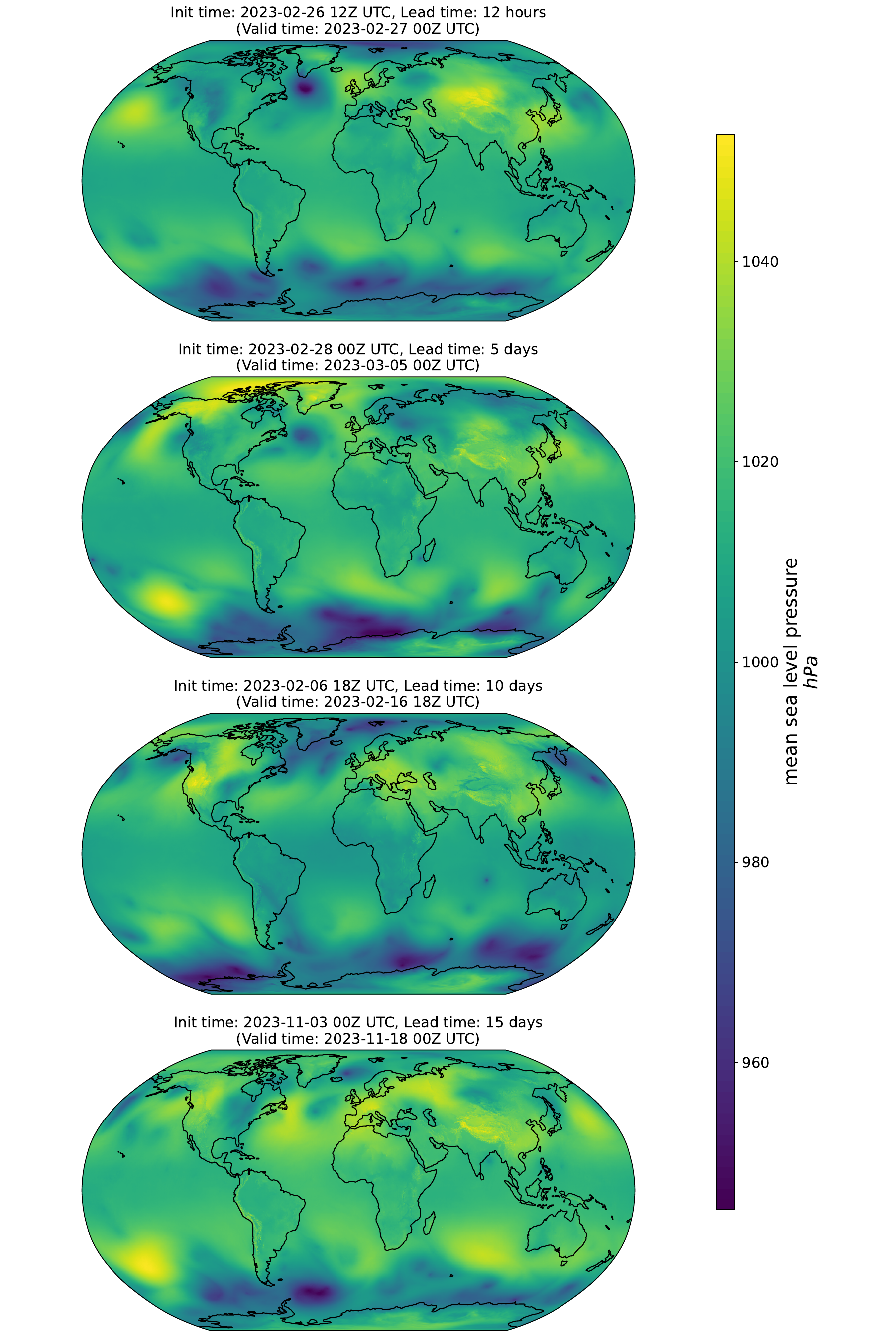}
    \caption{Visualization of mean sea level pressure.}
    \label{fig:sup:vis_msl}
\end{figure}

\begin{figure}
    \centering
    \includegraphics[width=0.9\textwidth]{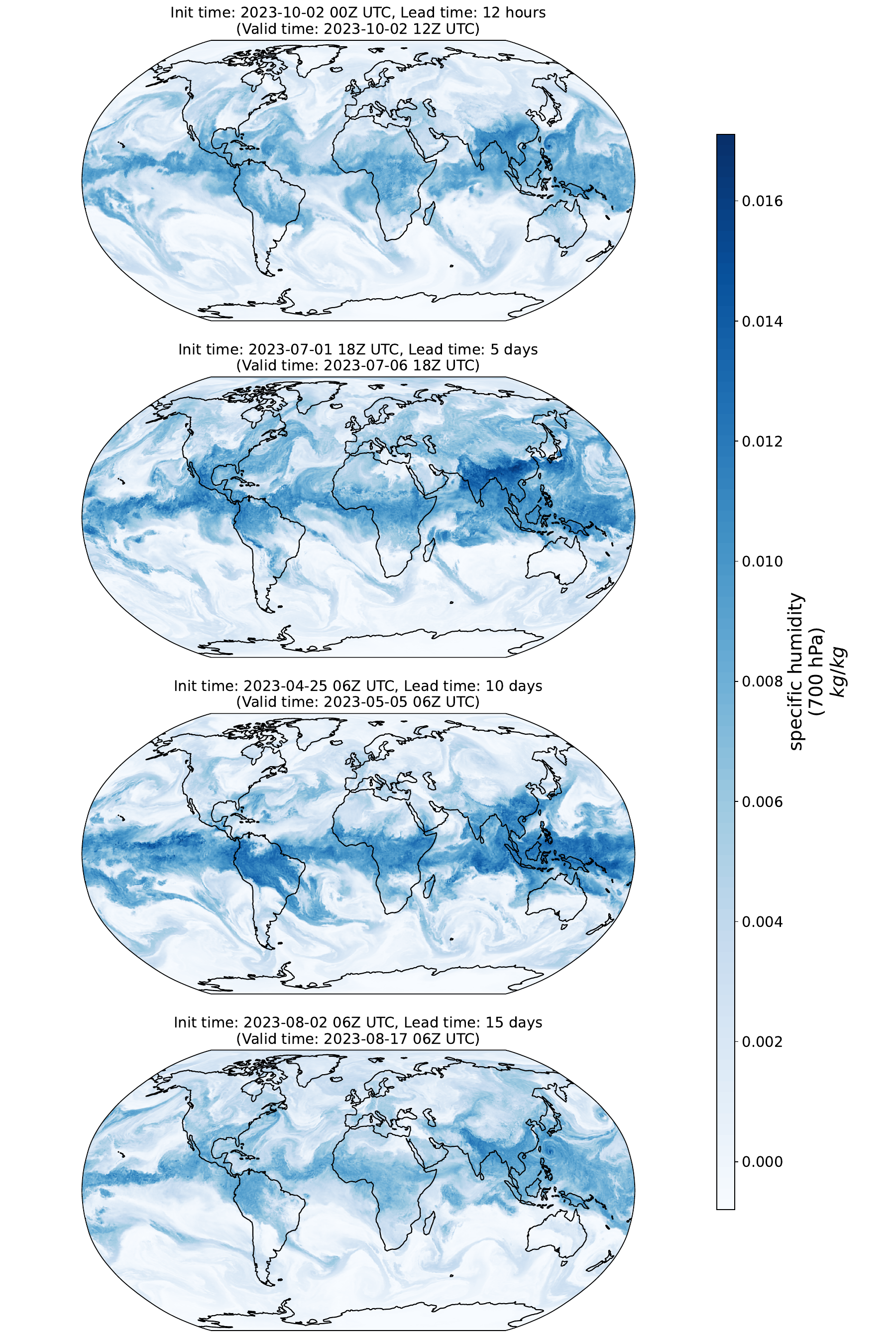}
    \caption{Visualization of specific humidity at 700 hPa.}
    \label{fig:sup:vis_q700}
\end{figure}

\begin{figure}
    \centering
    \includegraphics[width=0.9\textwidth]{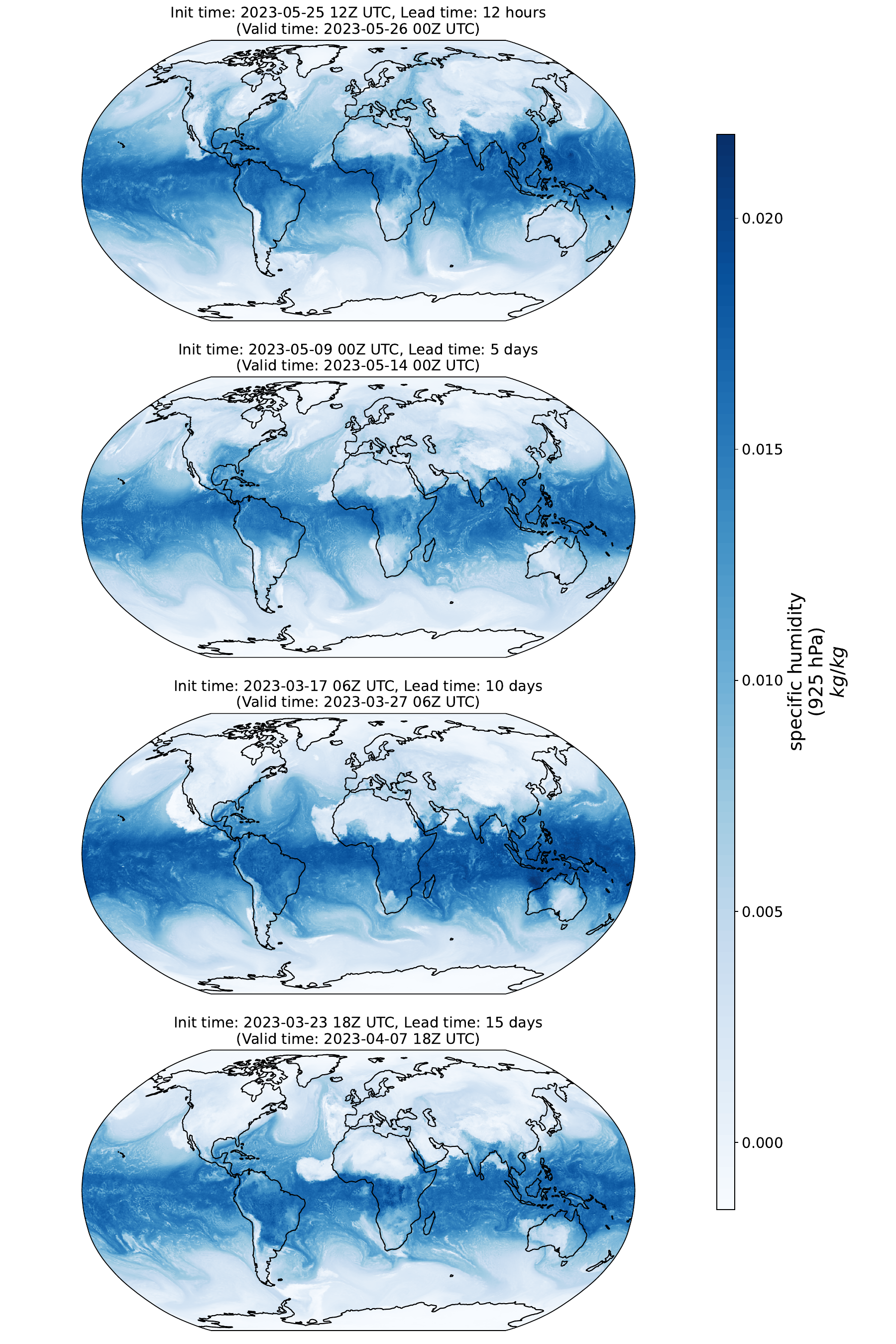}
    \caption{Visualization of specific humidity at 925 hPa.}
    \label{fig:sup:vis_q925}
\end{figure}

\begin{figure}
    \centering
    \includegraphics[width=0.9\textwidth]{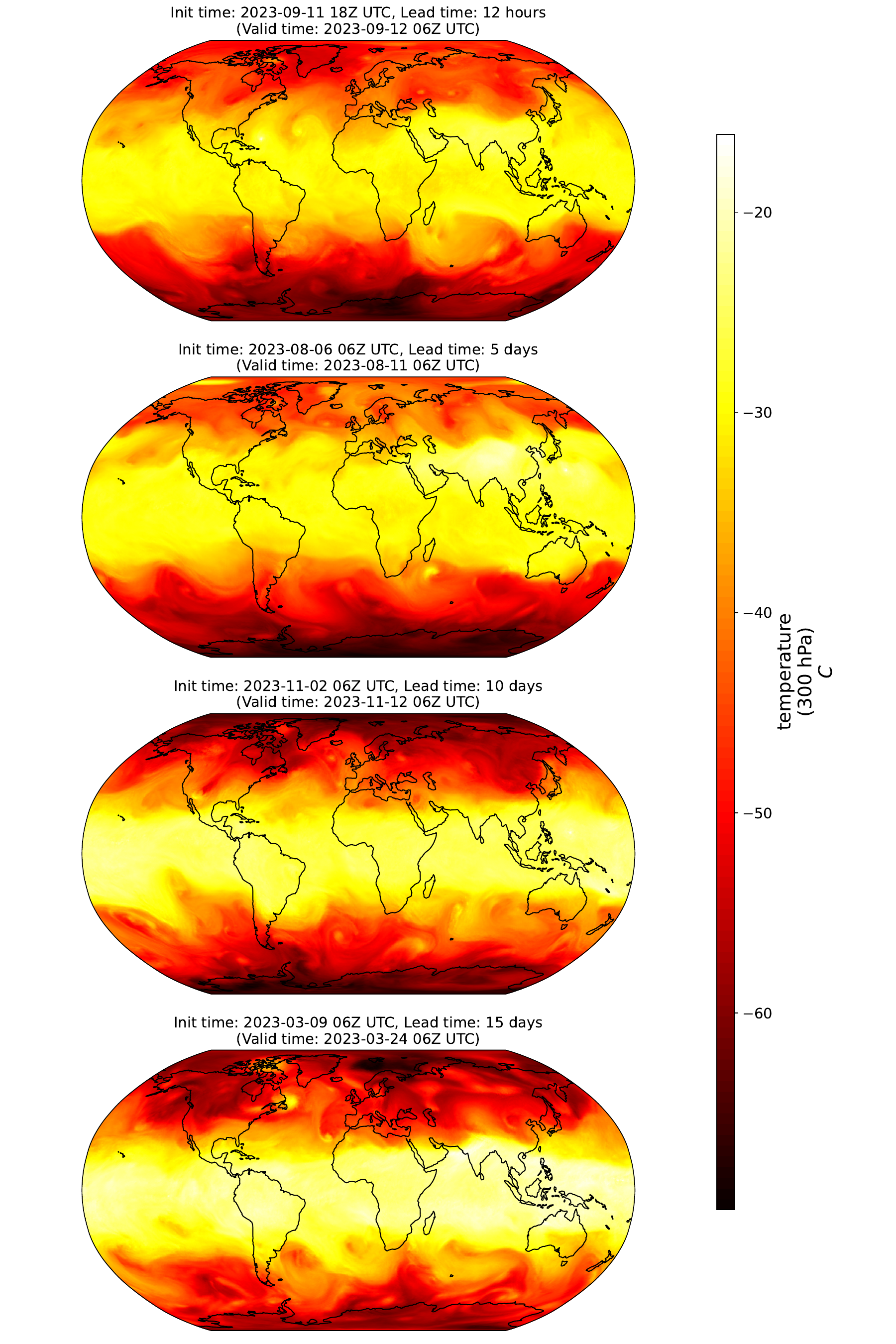}
    \caption{Visualization of temperature at 300 hPa.}
    \label{fig:sup:vis_t300}
\end{figure}

\begin{figure}
    \centering
    \includegraphics[width=0.9\textwidth]{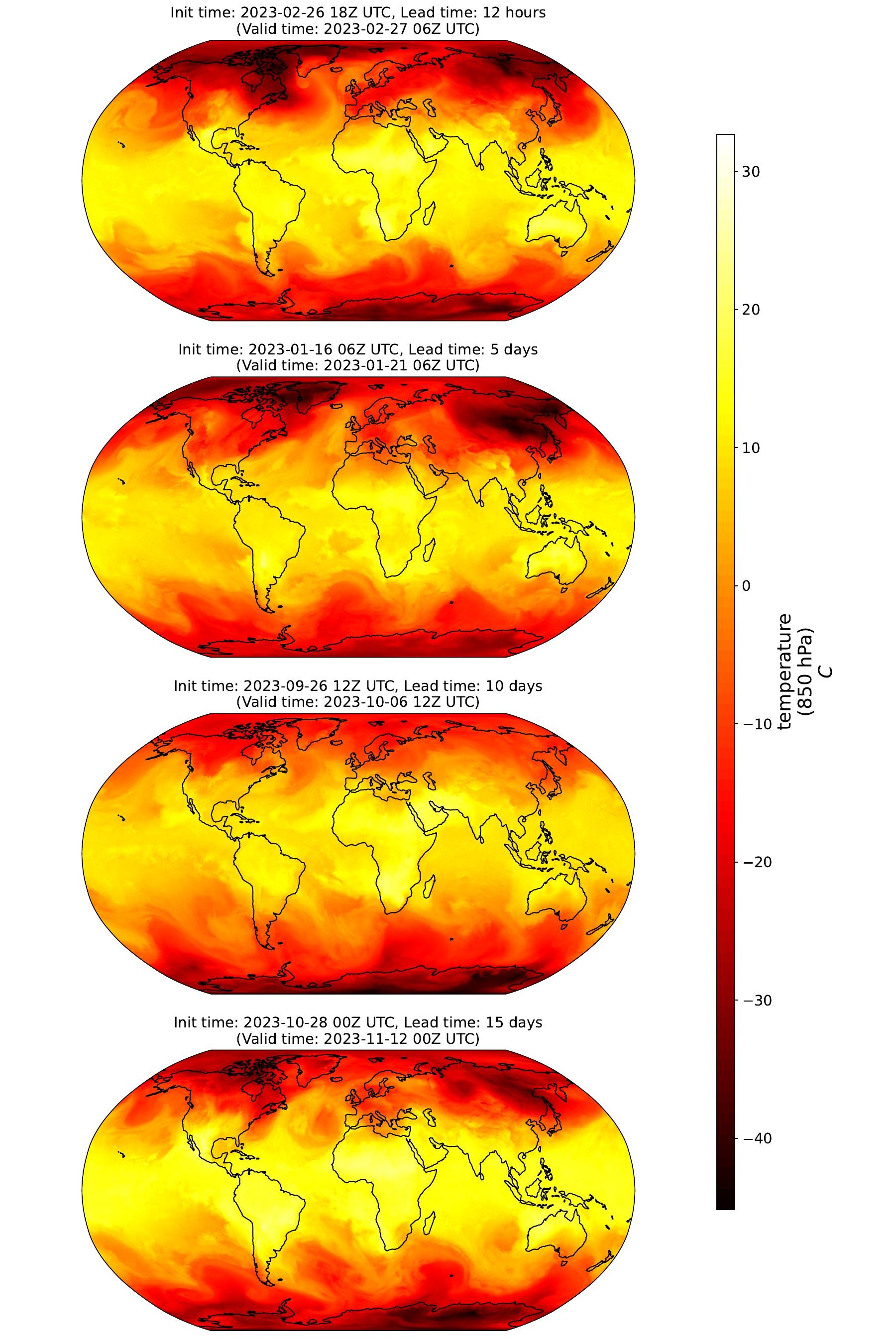}
    \caption{Visualization of temperature at 850 hPa.}
    \label{fig:sup:vis_t850}
\end{figure}

\begin{figure}
    \centering
    \includegraphics[width=0.9\textwidth]{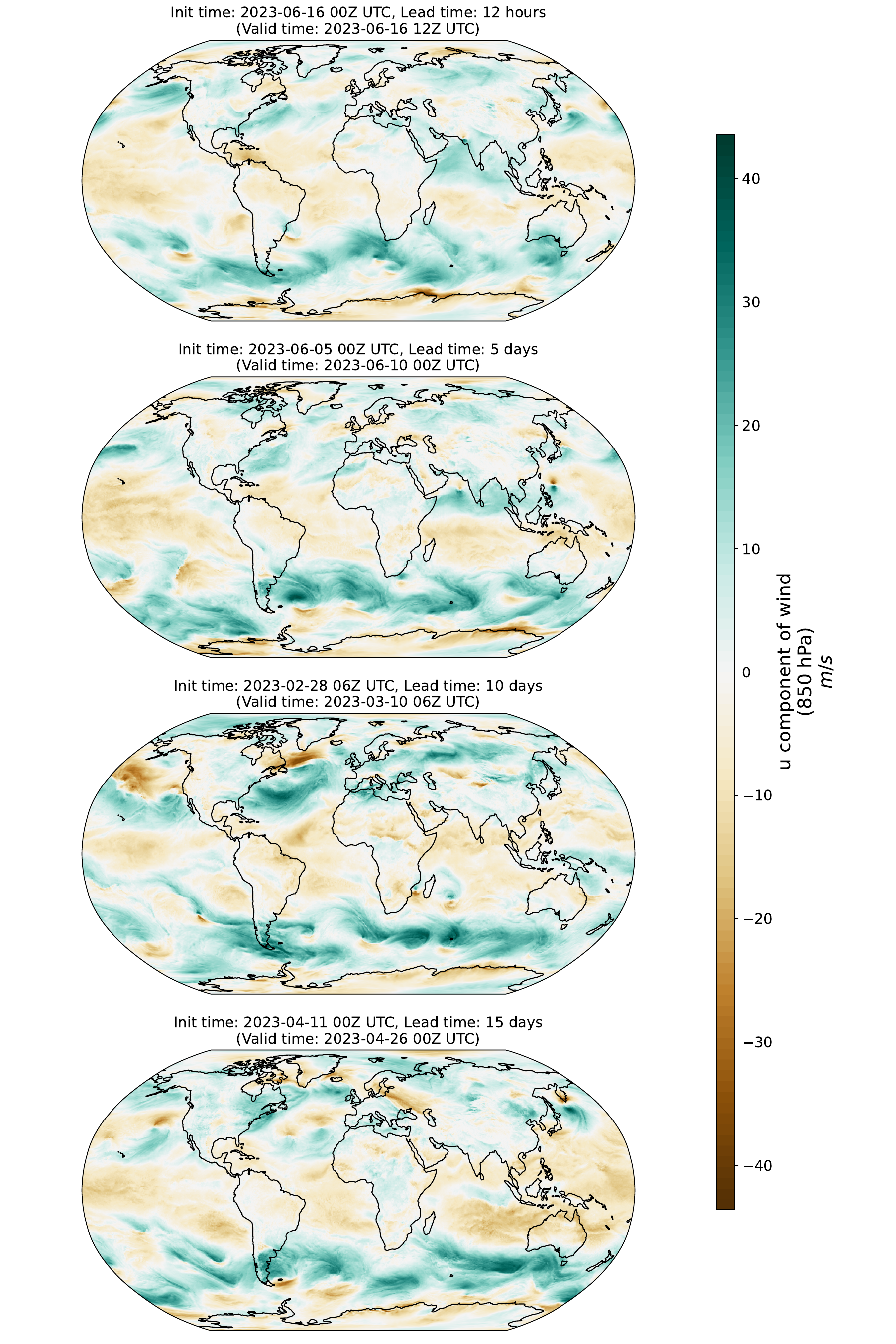}
    \caption{Visualization of u component of wind at 850 hPa.}
    \label{fig:sup:vis_u850}
\end{figure}

\begin{figure}
    \centering
    \includegraphics[width=0.9\textwidth]{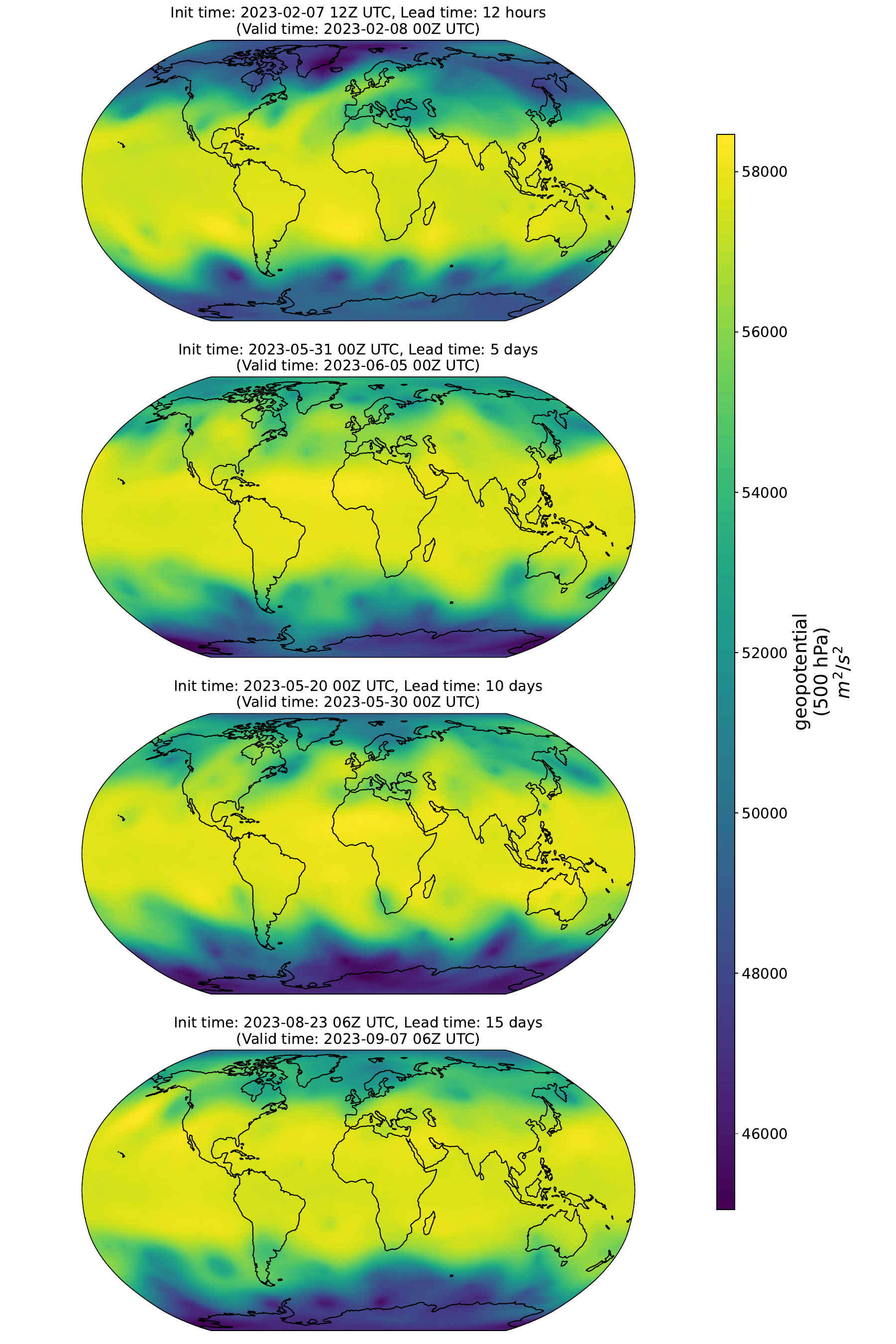}
    \caption{Visualization of geopotential at 500 hPa.}
    \label{fig:sup:vis_z500}
\end{figure}

\begin{figure}
    \centering
    \includegraphics[width=0.9\textwidth]{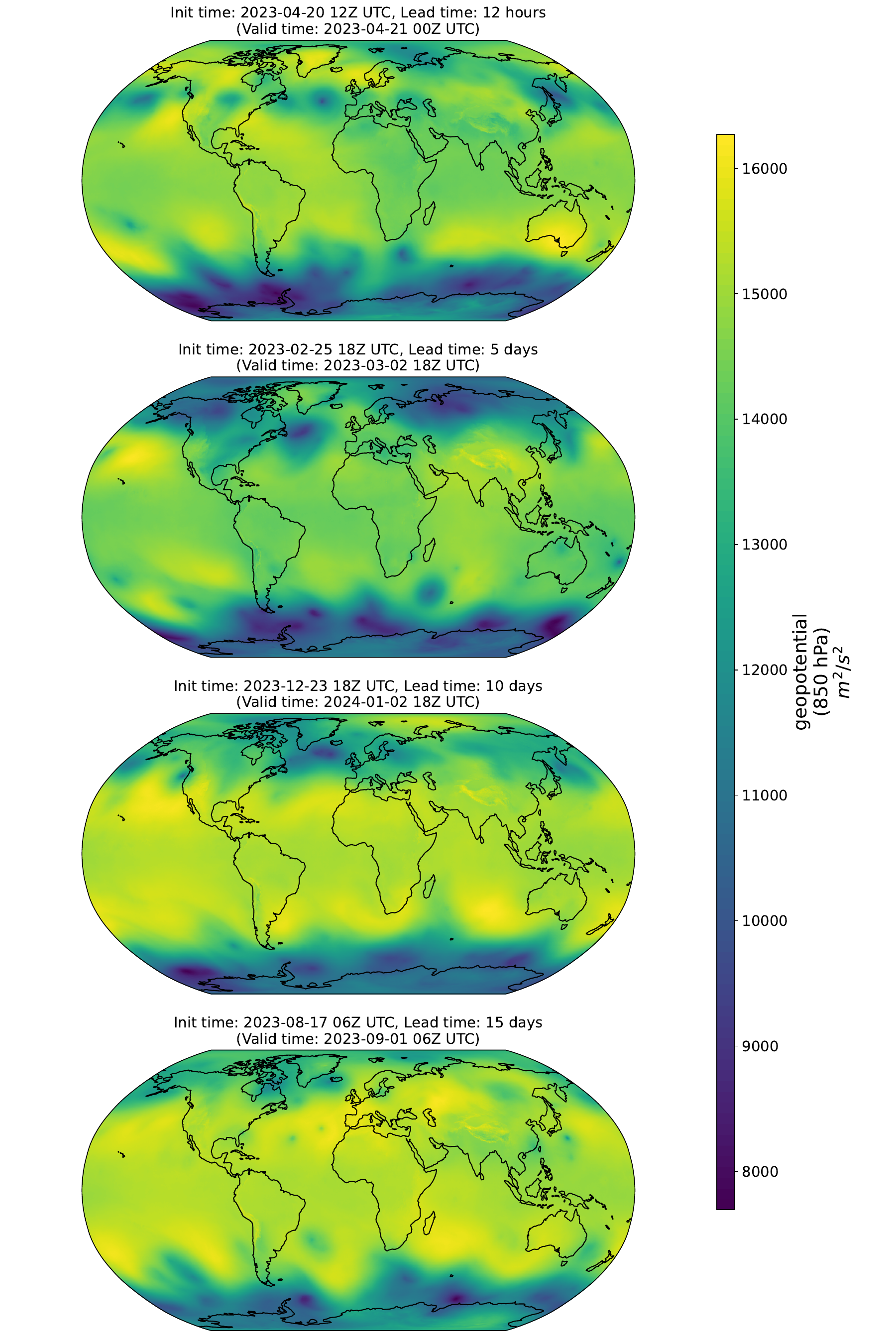}
    \caption{Visualization of geopotential at 850 hPa.}
    \label{fig:sup:vis_z850}
\end{figure}

We provide forecast visualizations for 10 representative variables: 2t (\cref{fig:sup:vis_2t}), 10u (\cref{fig:sup:vis_10u}, msl \cref{fig:sup:vis_msl}), q700 (\cref{fig:sup:vis_q700}), q925 (\cref{fig:sup:vis_q925}), t300 (\cref{fig:sup:vis_t300}), t850 (\cref{fig:sup:vis_t850}), u850 (\cref{fig:sup:vis_u850}), z500 (\cref{fig:sup:vis_z500}) and z850 (\cref{fig:sup:vis_z850}). For each variable-lead time combination we choose the 2023 initialization time with median CRPS error. This means that each figure show snapshots from a different forecast trajectory at each lead time. In each case we plot the first ensemble member.

\FloatBarrier

\end{document}